\documentclass[10pt]{article} 
\usepackage[accepted]{template/tmlr}

\usepackage{amsmath,amsfonts,bm}

\def\eqref#1{equation~\ref{#1}}

\def\1{\bm{1}}

\DeclareMathAlphabet{\mathsfit}{\encodingdefault}{\sfdefault}{m}{sl}
\SetMathAlphabet{\mathsfit}{bold}{\encodingdefault}{\sfdefault}{bx}{n}

\usepackage[most]{tcolorbox}
\usepackage{xcolor}
\usepackage{pifont}

\definecolor{surveyGreen}{HTML}{3F9D64}
\definecolor{surveyGreenDark}{HTML}{287A49}
\definecolor{surveyGreenLight}{HTML}{EAF7EE}
\definecolor{surveyGray}{HTML}{6F6F6F}
\definecolor{surveyGrayLight}{HTML}{F6F5F2}
\definecolor{surveyGrayBorder}{HTML}{D8D5CE}

\newtcolorbox{neutralheader}{
  enhanced,
  colback=white,
  colframe=surveyGrayBorder,
  boxrule=0.6pt,
  arc=1.5mm,
  left=2.2mm,
  right=2.2mm,
  top=1.5mm,
  bottom=1.5mm,
  height=1.45cm,
  valign=center
}

\newtcolorbox{ourheader}{
  enhanced,
  colback=surveyGreen,
  colframe=surveyGreenDark,
  boxrule=0.7pt,
  arc=1.5mm,
  left=2.2mm,
  right=2.2mm,
  top=1.5mm,
  bottom=1.5mm,
  height=1.45cm,
  valign=center
}

\newcommand{\neutralrow}[2]{%
  \begin{tcolorbox}[
    enhanced,
    colback=surveyGrayLight,
    colframe=surveyGrayBorder,
    boxrule=0.5pt,
    arc=1.3mm,
    left=2.2mm,
    right=2.2mm,
    top=1.3mm,
    bottom=1.0mm,
    height=1.40cm,
    valign=center,
    before skip=1.5mm
  ]
  {\color{surveyGray} #1}\par
  \vspace{0.4mm}
  {\footnotesize\bfseries #2}
  \end{tcolorbox}%
}

\newcommand{\ourrow}[2]{%
  \begin{tcolorbox}[
    enhanced,
    colback=surveyGreenLight,
    colframe=surveyGreen,
    boxrule=0.65pt,
    arc=1.3mm,
    left=2.2mm,
    right=2.2mm,
    top=1.3mm,
    bottom=1.0mm,
    height=1.40cm,
    valign=center,
    before skip=1.5mm
  ]
  {\color{surveyGreenDark} #1}\par
  \vspace{0.4mm}
  {\footnotesize\bfseries #2}
  \end{tcolorbox}%
}

\usepackage{template/packages}

\title{A Survey of Post-Training in Time Series Foundation Models}

\author{
\name Shifeng Xie\textsuperscript{*}
\email xidianxieshifeng@gmail.com \\
\addr Noah's Ark Lab; Universit\'e Paris Cit\'e
\AND
\name Ambroise Odonnat\textsuperscript{*}
\email ambroise.odonnat@gmail.com \\
\addr Noah's Ark Lab; Inria
\AND
\name Zehao Xiao
\email zxiao4ai@gmail.com \\
\addr Noah's Ark Lab
\AND
\name Lei Zan
\email xdzanlei@gmail.com \\
\addr Noah's Ark Lab
\AND
\name Malik Tiomoko
\email malik.tiomoko@huawei.com \\
\addr Noah's Ark Lab
\AND
\name Lujia Pan
\email panlujia@huawei.com \\
\addr Noah's Ark Lab
\AND
\name Themis Palpanas
\email themis@mi.parisdescartes.fr \\
\addr Universit\'e Paris Cit\'e
\AND
\name Boris N. Oreshkin\textsuperscript{\ensuremath{\dagger}}
\email oreshkin@amazon.com \\
\addr Amazon
\AND
\name Chenghao Liu\textsuperscript{\ensuremath{\ddagger}}
\email twinsken@gmail.com \\
\addr Datadog
\AND
\name Keli Zhang
\email zhangkeli1@huawei.com \\
\addr Noah's Ark Lab
}
\def\month{11}
\def\year{2026}
\def\openreview{\url{https://openreview.net/forum?id=LzPNfS7gWh}}
\begin{document}
\maketitle

\renewcommand{\thefootnote}{\fnsymbol{footnote}}

\footnotetext[1]{%
Equal contribution.
Correspondence to
\href{mailto:xidianxieshifeng@gmail.com}
{xidianxieshifeng@gmail.com}
and
\href{mailto:ambroise.odonnat@gmail.com}
{ambroise.odonnat@gmail.com}.
}

\begingroup
\renewcommand{\thefootnote}{}
\footnotetext{%
\textsuperscript{\ensuremath{\dagger}}Work not related to the author's position at Amazon.\quad
\textsuperscript{\ensuremath{\ddagger}}Work not related to the author's position at Datadog.
}
\endgroup

\renewcommand{\thefootnote}{\arabic{footnote}}

\begin{abstract}
Time series foundation models (TSFMs) have emerged as general-purpose models for time series analysis, but pretraining alone is often insufficient for reliable downstream deployment. Bridging this gap requires further intervention to handle domain shift, task heterogeneity, limited supervision, and computational constraints, which motivates post-training as a broad class of methods to adapt, augment, compose, calibrate, or specialize pretrained TSFMs for downstream tasks. In this work, we analyze TSFM post-training methods based on their locus of intervention in the prediction pipeline, yielding five categories: parameter adaptation, context augmentation, model composition, output processing and uncertainty control, and compression and specialization. Within each category, we study main representative methods and discuss their current limitations. We further identify future directions toward controlled adaptation, reliable context construction, uncertainty-aware model composition, calibrated output processing, and deployment-aware specialization. Overall, by providing a unifying framework for the emerging TSFM post-training landscape, this work aims to support future research to navigate the design space between a pretrained TSFM and its reliable downstream deployment.
\end{abstract}

\section{Introduction}
Time series data record how real-world systems evolve, capturing operationally critical quantities such as prices, demand, physiological measurements, energy load, weather variables, industrial sensor readings, and computing-resource usage. They arise across finance~\citep{sonkavde2023stock, lai2018modeling, tsay2005analysis, tsay2010multivariate}, healthcare~\citep{cepulionis2016electro, penfold2013healthcare, chui2017disease, gnassounou2025leveraginggenerictimeseries}, economics~\citep{nerlove2014economic}, electricity consumption~\citep{electricity,zhu2023eforecaster,qin2023personalized}, weather forecasting~\citep{weather,chen2023prompt,pathak2022fourcastnet}, supply management and retail~\citep{li2022supply,bose2017retail}, industrial processes, observability, and monitoring~\citep{cohen2026toto,liu2017fault}, and cloud computing~\citep{zhang2021cloudrca,yang2023dcdetector}. The typical tasks considered include forecasting, classification, anomaly detection, imputation, representation learning, and event-oriented analysis such as segmentation and change-point detection. 

\paragraph{From specialized methods $\ldots$} A plethora of methods have been developed in the past decades to solve these tasks, from mathematical methods~\citep{sorjamaa2007methodology, chen2021hamiltonian} and statistical approaches such as AutoRegressive Integrated Moving Average~\citep[ARIMA]{stevenson2007arima, box1990arima, box1974forecasting} and seasonal naive~\citep{hyndman2021seasonal}, to machine learning algorithms such as Support Vector Machines~\citep[SVMs]{hearst1998svm} and Gradient Boosting Machines~\citep[GBMs]{friedman2001gbm}. Before the recent shift toward foundation models, most deep learning approaches to time series were developed as task-specific architectures trained to solve a particular task~\citep{casolaro2023survey}, ranging from fully connected, recurrent and convolutional neural networks~\citep{smyl2019esrnn, rangapuram2018deep, salinas2020deepar, oreshkin2020nbeats, fan2017multistep, sen2019tcn} to transformer-based models~\citep{nie2023patchtst,ilbert2024samformer,moirai,haoyi2021informer,liu2024itransformer,wu2021autoformer,zhou2022fedformer}. Much of this progress has come from training global models over large collections of time series, rather than fitting separate models to individual series. By exploiting shared structure across series, this cross-series learning paradigm has contributed to the success of top-performing methods in forecasting competitions such as M4 and M5~\citep{smyl2019esrnn, anderer2022hierarchical}, helping establish deep learning as a dominant paradigm for time series applications.
\begin{figure}[!t]
    \centering
    \includegraphics[width=\linewidth]{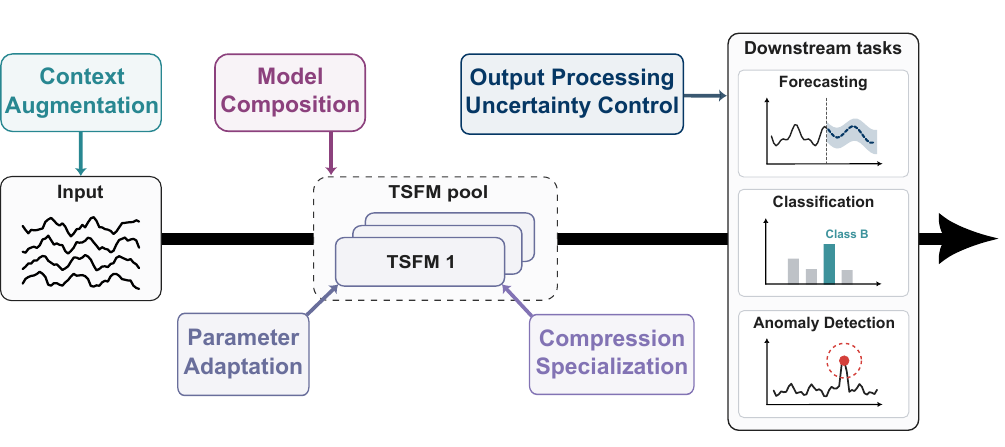}
    \caption{\textbf{Families of TSFMs post-training methods.} Many post-training methods transfer across tasks, domains, and pretrained architectures. As such, their main distinguishing feature is their locus of intervention in the prediction pipeline. This leads to the $5$ families of post-training methods studied in this work, which can act at the input context level, on the model parameters, or on the model outputs.}
    \label{fig:overview}
\end{figure}
\paragraph{$\ldots$ To foundation models.} Recently, the time series field has undergone the same paradigm shift as natural language processing~\citep{devlin2019bert,raffel2020t5,touvron2023llama,radford2018gpt,radford2019gpt2} and computer vision~\citep{caron2021dino, dosovitskiy2021vit,oquab2024dinov2,simeoni2025dinov3,he2022mae,kirillov2023sam,radford2021clip,carion2026sam,ravi2025sam}, moving from models trained on task-specific data towards pretraining on a large-scale corpora of diverse time series. Early transfer-learning work in time series showed that a model trained on source collections with a given sampling frequency could generalize directly to unseen target series without target-side adaptation~\citep{oreshkin2021meta}. The field then moved from frequency-specific transfer learning toward foundation-style pretraining, in which large sequence models are trained across diverse time series to support broad zero-shot forecasting. We refer to this emerging class of pretrained forecasting models as \emph{Time Series Foundation Models} (TSFMs). TimeGPT~\citep{timegpt1} was an early example of this direction, showing how pretrained transformer architectures could be adapted to forecasting. We note that several specific challenges occur when developing TSFMs, especially when it comes to building pretraining databases. Indeed, time series data are inherently noisy and often require expert knowledge to ensure their quality. This limits the open-source availability of high-quality data equivalent to ImageNet~\citep{deng2009imagenet} for computer vision and C4~\citep{dodge2021c4,raffel2020t5} or FineWeb~\citep{penedo2024fineweb} for LLMs. A lot of effort has been put into alleviating these issues, both from industry and academia. As such, many TSFM recipes were proposed over the years~\citep{moirai,chronos,liu2025moiraimoe,liu2026moirai20timeseries,mantis,cohen2026toto,das2024timesfm,ekambaram2024ttm,ansari2025chronos2,feofanov2026turningts2vecfoundationmodel,xie2026tabbyopenpretrainingrecipe}, with a noticeable convergence towards the transformer architecture~\citep{vaswani2017attention}. Methods to build pretraining databases were also proposed, including synthetic data~\citep{cauker,oreshkin2026sarsim0,taga2024timepfn,dooley2023forecastpfn}, and benchmarks and evaluations were proposed in tandem to measure the improvement of those models~\citep{shchur2026fevbench,cohen2026toto,agtabular,agtimeseries,aksu2024gifteval,dau2019ucr,bagnall2018uea}. 

\paragraph{The advent of post-training.} Despite the impressive capabilities of foundation models, pretraining is rarely the final step for practical deployment. For instance, base large language models (LLMs) acquire general knowledge during pretraining~\citep{radford2019gpt2,chinchilla,zhou2023lima} and become helpful chatbots after post-training~\citep{lambert2026reinforcementlearninghumanfeedback, instructgpt, xiao2024modeladaptationtesttime}, where they can be given reasoning abilities and be aligned with human intent~\citep{deepseek_r1,touvron2023llama,qwen3,gemma3}. 
In the context of TSFMs, post-training serves three main purposes. 
The first one has to do with task alignment. While pretraining enables learning general and reusable temporal knowledge, most downstream tasks have specific characteristics, such as the granularity, seasonality, feature correlations, the number of classes, or the forecasting horizon~\citep{moirai,das2024timesfm,ekambaram2024ttm}. 
Second, efficiently solving the task requires specialization, which can be achieved during post-training by adapting the model to the specific constraints at hand. 
Last but not least, TSFMs can benefit from strategies inspired by LLMs such as Retrieval-Augmented
Generation ~\citep[RAG]{RAG4CTS,crossrag,lewis2020rag}, or in-context learning~\citep[ICL]{brown2020fewshot,faw2025incontext}. 
Post-training methods help boost performance in these areas, improving the generalization to unseen distributions. While post-training turns pretrained LLMs into aligned chatbots, the post-training of TSFMs transforms them from universal temporal learners to specialized and reliable experts. 

\paragraph{Related work.} Technical overviews are important to help organize the field and give an overview of the current trends, limitations, and open problems.
Recent papers on post-training methods provide valuable insights but focus on large language models~\citep{lambert2026reinforcementlearninghumanfeedback,tie2025surveyposttraininglargelanguage,kumar2025llmposttrainingdeepdive,pmlr-v306-qi26b} and vision-language models~\citep{lambert2026reinforcementlearninghumanfeedback,tie2025surveyposttraininglargelanguage}. With the increase of TSFMs post-training approaches and given the important differences in terms of data, end-goals, and challenges, a specific treatment is needed for TSFMs. Many studies on TSFMs focus on the task of forecasting~\citep{casolaro2023survey}, which is prevalent in most recent papers~\citep{moirai,ansari2025chronos2,cohen2026toto,liu2026moirai20timeseries}, on the pretraining phase~\citep{TSFMSurvey2,TSFMSurvey}, on the role of synthetic data~\citep{liu2025surveysynthetic}, or on the emergence of multimodal time series analysis~\citep{multimodaltimeseriesSurvey}. Yet, a clear focus on post-training methods remains elusive. Close to our work is~\citet{TSFMsurvey3}, which describes the evolution from pretraining to post-training in TSFMs. However, the analysis remains limited to finetuning approaches, thus overlooking other important post-training components. In contrast, this study provides a structured overview, with explicit scope (see
Scope and methodology, above), of post-training approaches for TSFMs, categorizing them into the $5$ families of~\cref{fig:overview}, according to their locus of intervention in the prediction pipeline. We leverage this perspective to provide the structured taxonomy displayed in~\cref{fig:tsfm_posttraining_taxonomy}. Our tailored approach allows us to go beyond forecasting and to discuss the current trends, limitations, and open problems in the literature to help improve. Table~\ref{tab:related-surveys} compares this survey against the two closest prior works along
the dimensions most relevant to a reader deciding which to consult. \citep{TSFMSurvey} is a
broad tutorial covering the full TSFM lifecycle, with pretraining and general modeling choices
as its primary focus; post-training is mentioned but not organized as a dedicated taxonomy.
\citep{TSFMsurvey3} is the closest in scope, explicitly framing its contribution around the
pretraining-to-post-training transition, but restricts its analysis to finetuning-based
adaptation. This survey covers five loci of intervention -- parameter adaptation, context
augmentation, model composition, output processing, and compression -- of which finetuning
(parameter adaptation) is one.

\begin{table}[t]
\centering
\setlength{\tabcolsep}{0pt}

\begin{minipage}[t]{0.315\textwidth}
  \begin{neutralheader}
    {\color{surveyGray}Lifecycle-oriented survey}\\[-0.2mm]
    {\bfseries \citep{TSFMSurvey}}
  \end{neutralheader}

  \neutralrow
    {Primary scope}
    {End-to-end TSFM lifecycle}

  \neutralrow
    {Post-training breadth}
    {Adaptation within TSFM}

  \neutralrow
    {Beyond parameter updates}
    {No post-training taxonomy}

  \neutralrow
    {Task coverage}
    {Primarily forecasting}
\end{minipage}
\hfill
\begin{minipage}[t]{0.315\textwidth}
  \begin{neutralheader}
   {\color{surveyGray}Transition-oriented survey}\\[-0.2mm]
    {\bfseries \citep{TSFMsurvey3}}
  \end{neutralheader}

  \neutralrow
    {Primary scope}
    {Pretraining to post-training}

  \neutralrow
    {Post-training breadth}
    {Finetuning-centered}

  \neutralrow
    {Beyond parameter updates}
    {Limited coverage}

  \neutralrow
    {Task coverage}
    {Primarily forecasting}
\end{minipage}
\hfill
\begin{minipage}[t]{0.315\textwidth}
  \begin{ourheader}
    {\color{white}Unified post-training\\ framework (Ours)}
  \end{ourheader}

  \ourrow
    {Primary scope}
    {Post-training}

  \ourrow
    {Post-training breadth}
    {Five families across pipeline}

  \ourrow
    {Beyond parameter updates}
    {Context, composition, output, compression}

  \ourrow
    {Task coverage}
    {Forecasting, classification, anomaly detection}
\end{minipage}

\caption{
Positioning of our survey relative to the two closest TSFM surveys.
}
\label{tab:related-surveys}
\end{table}

\paragraph{Contributions.} This technical overview aims to give readers a clear understanding of the development and evolution of post-training in TSFMs, along with the limitations and areas of improvement. Our key contributions can be summarized as follows:
\begin{enumerate}[leftmargin=*]
    \item Structured overview. We provide a structured study of TSFM post-training methods,
covering a broad set of peer-reviewed publications and preprints meeting the criteria in the Scope and methodology paragraph above.
    \item \textbf{Structured taxonomy.} We provide a taxonomy of TSFMs post-training methods that clarifies where the intervention occurs in the prediction pipeline, offering insights into the evolution and limitations of current techniques.
    \item \textbf{Vision for future research.} To help foster the development of next-generation methods, we identify critical open problems and propose actionable directions for future work.
\end{enumerate}

\paragraph{Scope and methodology.} This survey reviews TSFM post-training methods publicly
available up to April 2026, including works from major machine learning venues and arXiv. We
consider methods that introduce a specific post-training technique and provide empirical
evaluation; withdrawn submissions are included when they merit discussion and are marked as withdrawn in Table~\ref{tab:comparison}.
Across the five categories in Fig.~\ref{fig:tsfm_posttraining_taxonomy}, we discuss representative approaches
selected to provide broad topical coverage rather than an exhaustive listing.

\paragraph{Overview.} Our paper is organized as follows. In~\cref{sec:background}, we recall some background notions on TSFMs, how they are pretrained, and how they are post-trained. The taxonomy of post-training methods is introduced in~\cref{sec:taxonomy} and organized according to where the post-training interventions occur on the prediction pipeline. In the following sections, we discuss in detail the different families of methods, their development, and current limitations: the parameter adaptation methods are discussed in~\cref{sec:parameter_adaptatation}, the context augmentation methods are discussed in~\cref{sec:context_augmentation}, the strategies of model composition are introduced in~\cref{sec:model_composition}, the approaches of output processing and uncertainty control are presented in~\cref{sec:output_processing_uncertainty} and, finally, the compression and specialization methods are presented in~\cref{sec:compression_specialization}.
Finally, in~\cref{sec:future_directions}, we synthesize the main limitations of current methods and outline future research directions.

\section{Background}
\label{sec:background}

\paragraph{Time Series Foundation Models.}

Let \(\mathbf{x}_{1:T} (\mathbf{x}_1,\ldots,\mathbf{x}_T)\in
\mathbb{R}^{T\times C}\) denote a time series of length \(T\) with \(C\) channels. Depending on the downstream task, the target output \(\mathbf{y}\in\mathcal{Y}\) can be a future trajectory, a class label, an imputed sequence, an anomaly score, or a predictive distribution\citep{TSFMSurvey,TSFMSurvey2}.

A time series foundation model is a pretrained model
\[
    f_{\theta_0}: \mathcal{X}\rightarrow \mathcal{Y},
\]
where \(\theta_0\) denotes parameters learned from large-scale time series corpora. Unlike task-specific models trained for a single dataset, horizon, or application, TSFMs are designed to transfer temporal representations~\citep{mantis} or predictive capabilities~\citep{chronos} across diverse downstream tasks. This positions them as a general-purpose modeling paradigm for time series analysis.

\paragraph{Pretraining of TSFMs.} The goal of TSFM pretraining is to learn broadly reusable temporal knowledge from a pretraining distribution, which may consist of real-world time series \citep{UniTS}, synthetic time series\citep{cauker,TempoPFN}, a mixture of both, or other modal data. Given a pretraining corpus  \(\mathcal{D}^{\mathrm{pre}}\), the model parameters are learned by optimizing a generic pretraining objective:
\[\theta_0 = \arg\min_{\theta} \mathbb{E}_{(x,y) \sim \mathcal{D}_{pre}} \left[ \mathcal{L}_{pre}(f_\theta, x, y) \right].\]
The pretraining objective may be based on forecasting \citep{moirai}, reconstruction \citep{moment}, masked modeling \citep{timesbert}, contrastive learning \citep{nutime}, in-context learning \citep{ticfm,fang2026chorustictrainingfreemultivariatetime}, or sequence-to-sequence prediction \citep{lagllama}. Through this process, the model is expected to capture common temporal patterns such as trends, seasonality, periodicity, local dynamics, cross-channel dependencies, and distributional uncertainty.

After pretraining, the obtained model \(f_{\theta_0}\) can be used as a general-purpose backbone for different time series tasks. However, the pretraining distribution is usually not identical to the target deployment distribution. Target domains may differ in temporal resolution, noise level, covariates, channel structure, forecasting horizon, label space, or nonstationary behavior.

\textbf{Post-training of TSFMs.} Although TSFMs are pretrained for broad transfer, practical deployment often requires adaptation to a target domain, task, or operational constraint~\citep{multimodaltimeseriesSurvey}. The most familiar mechanism is finetuning, where a pretrained model is updated on target-domain data~\citep{financial_finetuning,wasserstein_ft}. However, TSFM adaptation after pretraining extends beyond parameter updates, also involving context construction, retrieval, model composition, output refinement, uncertainty calibration, and deployment-oriented compression. We refer to post-training as any procedure applied after pretraining to improve a TSFM on a target domain \(\mathcal{D}^{\mathrm{tar}}\). Given a pretrained model
\(f_{\theta_0}\), a target input \(\mathbf{x}_{1:T}\), and optional adaptation resources \(\mathcal{R}\). Post-training can be written in a general form as:
\[
    \hat{\mathbf{y}}
    =
    \mathcal{A}
    \left(
        f_{\theta_0},
        \mathbf{x}_{1:T},
        \mathcal{R}
    \right),
\]
where \(\mathcal{A}\) denotes the post-training mechanism. The resource set \(\mathcal{R}\) may include labeled target-domain samples, unlabeled streams, calibration data, retrieved examples, auxiliary covariates, external memory, model pools, or deployment constraints.

Post-training is broader than conventional finetuning. It may update all or part of the model parameters \citep{financial_finetuning,var_forecasting}, add lightweight trainable modules \citep{chronosx,msft}, modify the input context \citep{tato}, retrieve relevant examples \citep{tsrag}, combine multiple pretrained models \citep{zoocast}, refine the output \citep{refinebridge}, or compress the model for efficient deployment \citep{li2026distillingtimeseriesfoundation}. In this sense, post-training is not a single technique, but a family of adaptation strategies that improve the usefulness of pretrained TSFMs under target domain and deployment requirements. This paper studies TSFM post-training methods based on the locus of intervention in the prediction pipeline. The overall taxonomy is provided in the following section.

\paragraph{Operational scope.} We consider a pretrained model a TSFM if it is trained on time series and other modal data drawn from multiple source domains or datasets under a generic (non-domain-specific)
objective, and can be applied zero-shot to series unseen during pretraining without
per-dataset retraining. Models trained and evaluated on a single dataset or a single domain
are excluded, even if architecturally similar to TSFMs. A post-training method is included if at least one of its reported experiments applies it to a TSFM in this sense; methods evaluated only with task-specific models are mentioned at most as related techniques and are excluded from Table~\ref{tab:comparison}.

We consider a procedure post-training if it is applied after pretraining, modifies the
model's parameters, input context, model composition, output space, or deployed computational
structure relative to the frozen pretrained checkpoint $f_{\theta_0}$, and persists across
inference calls rather than being constructed anew for each query. Under this definition,
learnable retrieval and memory modules (\S5) are included because their parameters or stored
content persist and are optimized rather than assembled per-query; ephemeral prompt
construction with no learned or persisted component is excluded. Methods with both an
input-side and output-side component, such as $\delta$-Adapter (\S7.1), are discussed once,
in the subsection corresponding to their primary locus of intervention, with the secondary
component noted explicitly to avoid double-counting.

\section{Taxonomy of TSFMs Post-Training Methods}
\label{sec:taxonomy}
\begin{figure}[!h]
    \centering
    \resizebox{\textwidth}{!}{%
\begin{tikzpicture}[
    font=\normalsize,
    >=Latex,
    node distance=0.4cm and 1.2cm, 
    trunk/.style={line width=0.9pt, draw=#1, rounded corners=2pt},
    root/.style={rounded corners=10pt, fill=tsfm!70!black, text=white, align=center, font=\bfseries\large, minimum width=3.6cm, minimum height=1.3cm},
    cat/.style={rounded corners=10pt, text=white, align=center, font=\bfseries, minimum width=3.4cm, minimum height=0.95cm},
    box/.style={rounded corners=8pt, align=left, text width=12.5cm, font=\footnotesize, inner sep=6pt, anchor=west}
]

\node[root] (root) at (0,0) {TSFMs\\Post-Training};

\node[cat, fill=param!80!black]       (A) at (4.5, 10.0)   {Parameter\\Adaptation};
\node[cat, fill=context!80!black]     (B) at (4.5, 4.5)    {Context\\Augmentation};
\node[cat, fill=composition!80!black] (C) at (4.5, 0.0)    {Model\\Composition};
\node[cat, fill=output!80!black]      (D) at (4.5, -5.2)   {Output\\\& Uncertainty};
\node[cat, fill=compression!80!black] (E) at (4.5, -10.5)  {Compression\\\& Specialization};


\node[box, draw=param!80!black, fill=param!6] (A2) at (8.0, 10.0) {%
\textbf{Parameter-Efficient Adaptation} $\bullet$ 
\work{ChronosX}{chronosx},
\work{UniCA}{unica},
\work{CoRA-Cov}{cora_covariate},
\work{WindPrompt}{wind_power_prompt},
\work{CoRA}{cora},
\work{AdaPTS}{adapts},
\work{DualWeaver}{dualweaver},
\work{GenPrompt}{generalized_prompt_tuning},
\work{MixFT}{mixft},
\work{TRACE}{trace},
\work{STAR}{star},
\work{MSFT}{msft},
\work{ULoRA-MoE}{uncertainty_ft},
\work{F2A}{f2a},
\work{Decision-FT}{decision_focused_ft},
\work{FORMED}{formed},
\work{VaR-Forecast}{var_forecasting},
\work{RV-Forecast}{realized_volatility}
};

\node[box, draw=param!80!black, fill=param!6, above=of A2] (A1) {%
\textbf{Full FT \& Continual PT} $\bullet$ 
\work{VisionTS++}{visiontspp},
\work{Wass-FT}{wasserstein_ft},
\work{Fin-FT}{financial_finetuning},
\work{Med-FedFT}{medical_federated_finetuning}
};

\node[box, draw=param!80!black, fill=param!6, below=of A2] (A3) {%
\textbf{Test-Time \& Online Adaptation} $\bullet$ 
\work{ELF}{elf},
\work{TAFAS}{tafas}
};

\node[box, draw=context!80!black, fill=context!6] (B2) at (8.0, 4.5) {%
\textbf{Memory Augmentation} $\bullet$ 
\work{MEMTS}{memts},
\work{TS-Memory}{tsmemory},
\work{MOMEMTO}{momemto}
};

\node[box, draw=context!80!black, fill=context!6, above=of B2] (B1) {%
\textbf{Retrieval Augmentation} $\bullet$ 
\work{RAF}{raf},
\work{TimeRAF}{timeraf},
\work{TS-RAG}{tsrag},
\work{Cross-RAG}{crossrag},
\work{RATFM}{ratfm},
\work{RAG4CTS}{RAG4CTS}
};

\node[box, draw=context!80!black, fill=context!6, below=of B2] (B3) {%
\textbf{Context Transformation} $\bullet$ 
\work{TATO}{tato}
};

\node[box, draw=composition!80!black, fill=composition!6] (C2) at (8.0, 0.0) {%
\textbf{Adaptive Fusion} $\bullet$ 
\work{Synapse}{synapse},
\work{Boosting}{boosting}
};

\node[box, draw=composition!80!black, fill=composition!6, above=of C2] (C1) {%
\textbf{Model Selection} $\bullet$ 
\work{Chroma}{chroma},
\work{ZooCast}{zoocast},
\work{TimeTic}{timetic}
};

\node[box, draw=composition!80!black, fill=composition!6, below=of C2] (C3) {%
\textbf{Sequential \& Agentic Fusion} $\bullet$ 
\work{SeqFusion}{seqfusion},
\work{Conversational TSFM}{conversational}
};

\node[box, draw=output!55!black, fill=output!4] (D2) at (8.0, -4.5) {%
\textbf{Forecast Calibration} $\bullet$
\work{FM+CP}{fm_conformal_prediction}
};

\node[box, draw=output!55!black, fill=output!4, above=of D2] (D1) {%
\textbf{Forecast Refinement} $\bullet$ 
\work{\(\delta\)-Adapter}{delta_adapter},
\work{TFMAdapter}{tfmadapter},
\work{RefineBridge}{refinebridge}
};

\node[box, draw=output!55!black, fill=output!4, below=of D2] (D3) {%
\textbf{Probabilistic Output Modeling} $\bullet$ 
\work{Corr. Sample Paths}{correlated_sample_paths}
};

\node[box, draw=output!55!black, fill=output!4, below=of D3] (D4) {%
\textbf{Anomaly Detection} $\bullet$ 
\work{Adaptive Conformal AD}{adaptive_conformal},
\work{Complexity+Stats AD}{complexity_statistics}
};

\node[box, draw=compression!55!black, fill=compression!6] (E1) at (8.0, -9.8) {%
\textbf{Knowledge Distillation} $\bullet$ 
\work{DistilTS}{li2026distillingtimeseriesfoundation},
\work{Battery-Timer}{chan_2026_battery}
};

\node[box, draw=compression!55!black, fill=compression!6, below=of E1] (E2) {%
\textbf{Pruning \& Specialization} $\bullet$ 
\work{Less-is-More}{zhao2026less},
\work{Flow-of-Ranks}{yu2026understanding},
\work{Consensus-Subspace KD}{Consensus-Subspace}
};

\foreach \i/\col in {A/param, B/context, D/output, E/compression} {
    \draw[trunk=tsfm!50!black, -{Stealth[scale=0.5]}] (root.east) -- ++(0.7,0) |- (\i.west);
}
\draw[trunk=tsfm!50!black, -{Stealth[scale=0.5]}] (root.east) -- (C.west);

\foreach \cat/\boxes/\col in {
    A/{A1,A2,A3}/param,
    B/{B1,B2,B3}/context,
    C/{C1,C2,C3}/composition,
    D/{D1,D2,D3,D4}/output,
    E/{E1,E2}/compression%
} {
    \coordinate (junction) at ($(\cat.east)+(0.8,0)$);
    \draw[\col!80!black, line width=0.9pt] (\cat.east) -- (junction);
    \foreach \b in \boxes {
        \draw[-{Stealth[scale=0.6]}, \col!80!black, line width=0.9pt, rounded corners=2pt] (junction) |- (\b.west);
    }
}
\end{tikzpicture}%
}
    \caption{Taxonomy of post-training methods for time series foundation models. Methods are grouped by the locus of intervention: parameter space, input context, model composition, output processing, and computational structure. Each method is assigned to its primary subcategory to avoid duplication.
    }
    \label{fig:tsfm_posttraining_taxonomy}
\end{figure}
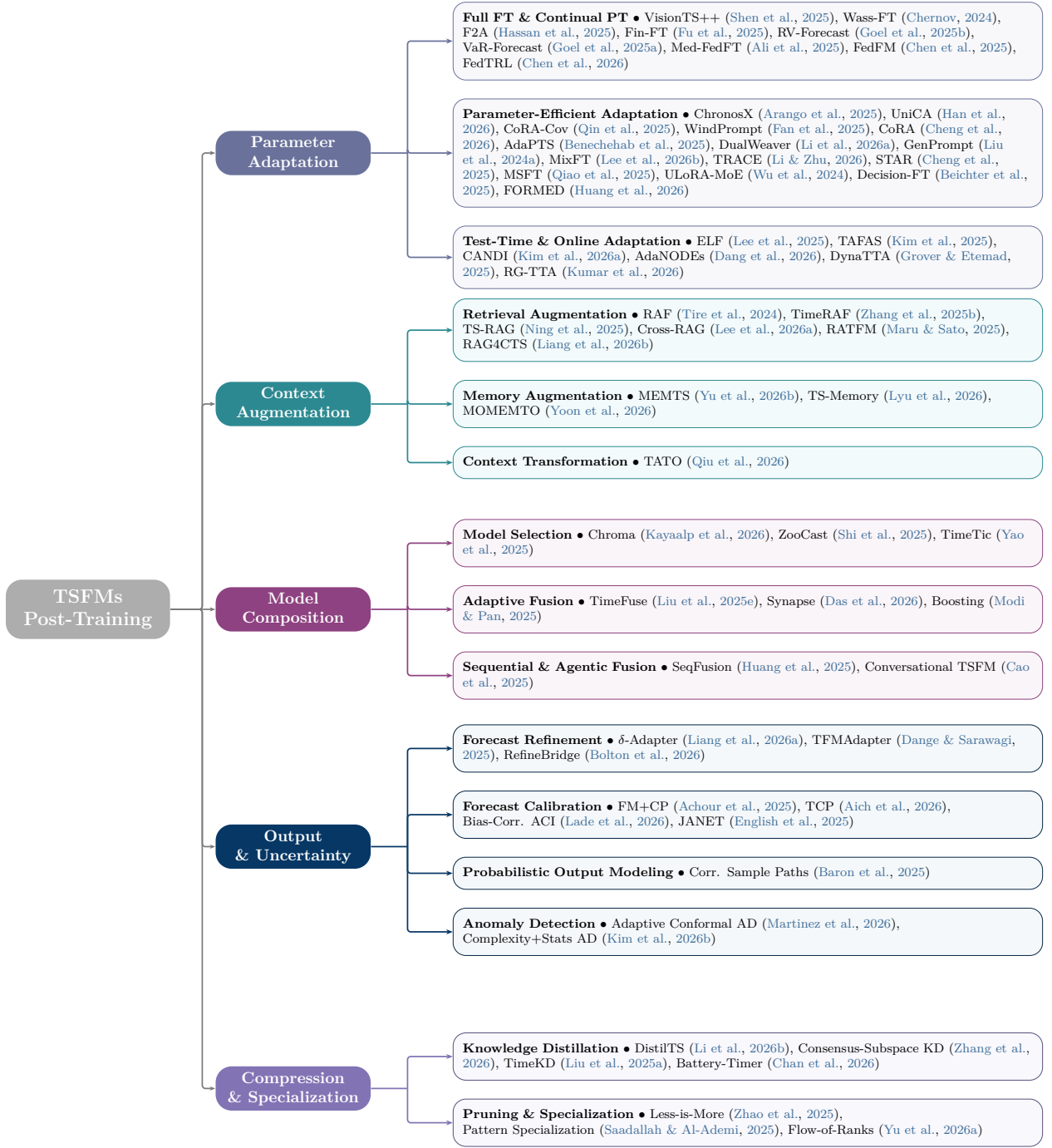

We organize existing TSFM post-training methods by their locus of intervention in the inference pipeline, rather than by downstream task, model backbone, or application domain. Given a pretrained time series foundation model
\(f_{\theta_0}: \mathcal{X}\rightarrow\mathcal{Y}\), we define post-training as any procedure that improves its performance, robustness, calibration, or deployment efficiency on a target domain \(\mathcal{D}_{\mathrm{tar}}\) by modifying one of these components. Since many of these methods transfer across tasks, domains, and pretrained backbones, what principally distinguishes them is where they intervene. This motivates a taxonomy of five categories, described in~\cref{fig:overview}: modifications in the model parameters, input context, model composition, output space, or computational structure. The overall taxonomy of the methods is provided in~\cref{fig:tsfm_posttraining_taxonomy}.

\paragraph{\color{param!90!black}A. Parameter adaptation.}
Parameter adaptation intervenes directly in the model parameters of a pretrained TSFM. It either updates some or all of the original model parameters, as in full-parameter finetuning, continual pretraining, and test-time adaptation, or introduces additional trainable components while keeping most of the backbone frozen, as in parameter-efficient finetuning. 
This can be formally expressed as the following parameter update:
\[
    \theta^\star = \theta_0 + \Delta\theta ,
\]
where \(\Delta\theta\) denotes the post-training update, ranging from full-model parameter changes to a small set of new trainable parameters.

\paragraph{\color{context!90!black}B. Context augmentation.}
Context augmentation intervenes on the input side of a TSFM with retrieved examples, external memory, support samples, or in-context augmentation. Instead of primarily changing the backbone parameters, these methods construct an additional context \(\mathcal{C}_x\) for each input series:
\[
    \hat{\mathbf{y}}_{1:H}
    =
    f_{\theta_0}\bigl(\mathbf{x}_{1:T},\mathcal{C}_x\bigr).
\]
We further categorize the methods into retrieval augmentation, memory augmentation, and in-context augmentation.

\paragraph{\color{composition!90!black}C. Model composition.}
Model composition treats TSFM post-training as a model allocation problem, combining multiple pretrained or adapted models rather than relying on a single one.
Given a pool of pretrained or adapted models: $\mathcal{M}=\{f_{\theta_1},\ldots,f_{\theta_K}\}$,
a router \(r_\phi\) assigns each target input to the models based on the input series and, optionally, model metadata such as their training domains, supported frequencies, context lengths, horizons, validation performance, or uncertainty estimates. A common soft-routing formulation is
\[
    \boldsymbol{\alpha}
    =
    r_\phi(\mathbf{x}_{1:T}, \mathcal{M})
    \in \Delta^{K-1},
    \qquad
    \hat{\mathbf{y}}_{1:H}
    =
    \sum_{k=1}^{K}
    \alpha_k f_{\theta_k}(\mathbf{x}_{1:T}),
\]
where \(\Delta^{K-1}
=
\{\boldsymbol{\alpha}\in\mathbb{R}_{+}^{K}:\sum_{k=1}^{K}\alpha_k=1\}\)
is the probability simplex over the \(K\) models. The router parameters \(\phi\) may be learned from validation or target-domain data, estimated from historical model performance, induced by retrieval or similarity scores, or specified by metadata-based rules using properties such as domain, frequency, horizon, context length, or uncertainty. This category includes model selection, ensembling, fusion, routing, and mixture-of-experts-style composition.

\paragraph{\color{output!90!black}D. Output processing and uncertainty control.}
Output processing and uncertainty control intervene on the initial output of a pretrained TSFM $\tilde{\mathbf{y}}_{1:H} = f_{\theta_0}(\mathbf{x}_{1:T})$. An output-level post-processor \(g_\psi\) then maps the raw output, possibly together with the input context, to a refined prediction, calibrated interval, probabilistic sample, or anomaly score:
\[
    \hat{\mathbf{y}}_{1:H}
    =
    g_\psi
    \left(
    \tilde{\mathbf{y}}_{1:H},
    \mathbf{x}_{1:T}
    \right).
\]
The parameters \(\psi\in\Psi\) can be learned from target-domain residuals, estimated on a calibration set, updated online from recent forecast errors, or set by distributional or conformal methods. Depending on the objective, \(\psi\) can include correction-model weights, residual quantiles, conformal thresholds, variance or scale parameters, recalibration maps, or anomaly-score normalization constants. Thus, this category encompasses output refinement, conformal calibration, probabilistic recalibration, residual correction, and anomaly-score processing techniques.

\paragraph{\color{compression!90!black}E. Compression and specialization.}
Compression and specialization operate during the deployment of pretrained TSFM. The goal is to convert a broadly pretrained model \(f_{\theta_0}\) into a cheaper, faster, or more targeted variant \(f_{\bar{\theta}}\) for a specific deployment regime, such that:
\[
    \mathrm{Cost}(f_{\bar{\theta}})
    <
    \mathrm{Cost}(f_{\theta_0}),
\]
while maintaining competitive accuracy, calibration, and robustness on the target domain. Here, \(\mathrm{Cost}\) can quantify parameter count, memory footprint, inference latency, context-length complexity, energy consumption, or serving cost. The specialized parameters \(\bar{\theta}\) can be obtained through techniques such as distillation, pruning, quantization, low-rank approximation, architecture replacement, or targeted adaptation to a restricted set of domains, horizons, or tasks. This category includes both model compression for efficient deployment and specialization for constrained operational settings.

\subsection{Practical Selection Guidance}
\begin{figure}[t]
\centering
\begin{tikzpicture}[
  node distance=0.9cm and 1.9cm,
  >={Stealth[length=2.3mm,width=1.8mm]},
  every node/.style={font=\small},
  box/.style={
    draw=steel, rounded corners=3pt, align=center,
    minimum width=3.4cm, minimum height=0.95cm,
    fill=steel, text=white, font=\small\bfseries,
    drop shadow={shadow xshift=0.6mm,shadow yshift=-0.6mm,opacity=0.25}
  },
  leaf/.style={
    draw=teal!70!black, rounded corners=3pt, align=center,
    minimum width=2.9cm, minimum height=0.95cm,
    fill=teal!12, text=teal!45!black, font=\small\bfseries,
    drop shadow={shadow xshift=0.6mm,shadow yshift=-0.6mm,opacity=0.2}
  },
  lbl/.style={
    draw=amber!80!black, fill=amber!15, rounded corners=6pt,
    inner sep=1.5pt, font=\scriptsize\bfseries, text=amber!50!black
  },
  edgeline/.style={draw=steel, line width=0.9pt}
]

\node[box] (start) {Select post-training method};
\node[box, below=of start] (privacy) {Target data leave site?};
\node[leaf, right=of privacy] (fed) {Federated FT\\\scriptsize(\S4.1)};
\node[box, below=of privacy] (volume) {Target data volume};
\node[leaf, right=of volume] (context) {Context aug.\\\scriptsize(\S5)};
\node[box, below=of volume] (latency) {Tight inference budget?};
\node[leaf, right=of latency] (compress) {Compress / PEFT\\\scriptsize(\S4.2, \S8)};
\node[leaf, below=of latency] (full) {Full FT / fusion\\\scriptsize(\S4.1, \S6)};

\draw[edgeline,->] (start) -- (privacy);
\draw[edgeline,->] (privacy) -- (volume) node[lbl,midway,right=1pt] {Yes};
\draw[edgeline,->] (privacy) -- (fed)    node[lbl,midway,above=1pt] {No};
\draw[edgeline,->] (volume)  -- (latency) node[lbl,midway,right=1pt] {Abundant};
\draw[edgeline,->] (volume)  -- (context) node[lbl,midway,above=1pt] {Little/none};
\draw[edgeline,->] (latency) -- (full)     node[lbl,midway,right=1pt] {No};
\draw[edgeline,->] (latency) -- (compress) node[lbl,midway,above=1pt] {Yes};

\end{tikzpicture}
\caption{Decision tree for selecting a TSFM post-training family under deployment constraints. Output processing (D) is composable with any branch and is omitted here for clarity; see Table~\ref{tab:selection-guidance}.}
\label{fig:decision-tree}
\end{figure}
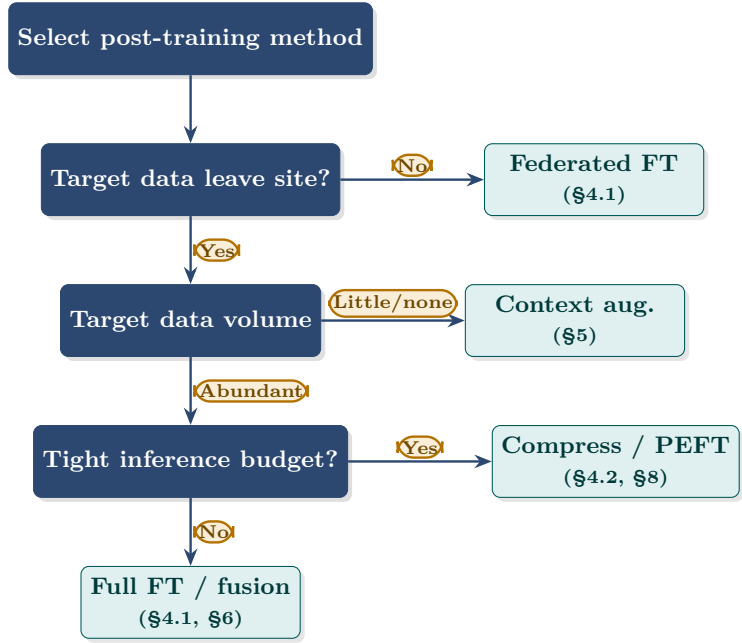

While the following sections discuss each family in depth, we summarize here how the constraints most
relevant to deployment---target-data volume, compute/latency budget, and privacy---map onto the
five categories of Fig.~\ref{fig:tsfm_posttraining_taxonomy}. Table~\ref{tab:selection-guidance} and
Fig.~\ref{fig:decision-tree} are grounded entirely in evidence already discussed in the
per-category sections. Output processing (D) and compression (E) are largely orthogonal to this
choice: D is a lightweight calibration layer composable on top of any of A--C, and E is commonly
applied \emph{after} a specialization decision has already been made via A or C (\S8.3).

\begin{table}[t]
\centering
\small
\renewcommand{\arraystretch}{1.5}
\setlength{\tabcolsep}{7pt}
\begin{tabular}{
  >{\raggedright\arraybackslash}p{2.3cm}
  >{\raggedright\arraybackslash}p{1.7cm}
  >{\raggedright\arraybackslash}p{2.7cm}
  >{\centering\arraybackslash}p{1.6cm}
  >{\raggedright\arraybackslash}p{3.7cm}
}
\toprule
\textbf{\textcolor{black}{Category}} & \textbf{\textcolor{black}{Target-data need}} & \textbf{\textcolor{black}{Compute/latency cost}} & \textbf{\textcolor{black}{Privacy fit}} & \textbf{\textcolor{black}{When to prefer}} \\
\midrule

\rowcolor{catA!8}
\textcolor{catA}{\textbf{A1}}~Full FT / continual PT & Moderate--large & High offline; none at inference & \textcolor{red!70!black}{\textbf{Poor}}\textsuperscript{*} & Severe shift, data available (\S4.4) \\
\rowcolor{catA!8}
\textcolor{catA}{\textbf{A2}}~PEFT & Small--moderate & Low training; frozen backbone & \textcolor{gray!60!black}{\textbf{Neutral}} & Mild shift, multi-task sharing (\S4.2) \\
\rowcolor{catA!8}
\textcolor{catA}{\textbf{A3}}~Test-time / online & None & Low per-step, runs online & \textcolor{gray!60!black}{\textbf{Neutral}} & Nonstationary streams (\S4.3) \\

\midrule
\rowcolor{catB!8}
\textcolor{catB}{\textbf{B1}}~Retrieval aug. & None (KB needed) & Latency/index overhead & \textcolor{orange!85!black}{\textbf{Depends}} & Relevant history exists (\S5.4) \\
\rowcolor{catB!8}
\textcolor{catB}{\textbf{B2}}~Memory aug. & None & Lower latency than B1 & \textcolor{catE}{\textbf{Better}} & B1 too slow (\S5.2) \\
\rowcolor{catB!8}
\textcolor{catB}{\textbf{B3}}~Context transform. & None & Very low & \textcolor{catE}{\textbf{Good}} & Input-only mismatch (\S5.3) \\

\midrule
\rowcolor{catC!8}
\textcolor{catC}{\textbf{C}}~Model composition & Validation data & Selection low; fusion higher & \textcolor{gray!60!black}{\textbf{Neutral}} & Complementary models exist (\S6.4) \\

\midrule
\rowcolor{catD!8}
\textcolor{catD}{\textbf{D}}~Output processing & Small calib.\ set & Very low, post-hoc & \textcolor{catE}{\textbf{Good}} & Composable add-on (\S7.5) \\

\midrule
\rowcolor{catE!8}
\textcolor{catE}{\textbf{E}}~Compression & Teacher + some data & High one-time; large savings & \textcolor{gray!60!black}{\textbf{Neutral}} & Edge/high-throughput deploy (\S8.2) \\

\bottomrule
\end{tabular}
\caption{Post-training method selection by deployment constraint. Color-coded categories
follow Fig.~\ref{fig:tsfm_posttraining_taxonomy}; privacy-fit ratings are colored for quick
scanning (\textcolor{red!70!black}{\textbf{Poor}}~$\rightarrow$~\textcolor{gray!60!black}{\textbf{Neutral}}~$\rightarrow$~\textcolor{catE}{\textbf{Good/Better}}).
\textsuperscript{*}Poor unless federated (\S4.1).}
\label{tab:selection-guidance}
\end{table}

\subsection{Interactions Between Post-training Families}
Sections~4--8 discuss each family in isolation, but in practice post-training methods are
routinely combined rather than used as mutually exclusive alternatives. Table~\ref{tab:interactions}
summarizes how the five families interact pairwise, distinguishing methods that compose cleanly
from those that require additional care because one method changes an assumption the other relies on.

A central example is uncertainty control (D) after parameter adaptation (A): forecast calibration
and conformal methods (\S7.2) are typically fit on a calibration set assuming a fixed underlying
model. Test-time and online adaptation (\S4.3), by construction, updates the model \emph{during}
deployment, so a calibration set collected before adaptation no longer reflects the post-adaptation
error distribution, and coverage guarantees can silently degrade. The same issue arises, to a lesser
extent, whenever retrieved or transformed context (B) changes the input distribution seen by an
output-level calibrator (D). This motivates recalibration or online calibration update as a
dependency of, rather than an alternative to, test-time adaptation and context augmentation ---
an interaction the current taxonomy does not surface because each family is discussed independently.

Conversely, some interactions are already exploited productively in the literature we survey:
compression and specialization (E) are frequently applied \emph{after} a parameter-adaptation or
model-composition decision has fixed a specialized model to compress (\S8.3), and model composition
(C) can route across a pool of independently PEFT-adapted models (\S4.2, \S6.1) rather than treating
adaptation and composition as competing choices.

\begin{table}[t]
\centering
\small
\renewcommand{\arraystretch}{2.0}
\setlength{\tabcolsep}{5pt}
\begin{tabular}{
  >{\raggedright\arraybackslash}p{1.8cm}
  >{\raggedright\arraybackslash}p{2.7cm}
  >{\raggedright\arraybackslash}p{2.7cm}
  >{\raggedright\arraybackslash}p{2.7cm}
  >{\raggedright\arraybackslash}p{2.7cm}
}
\toprule
\cellcolor{white} & \cellcolor{catB!15}\textbf{\textcolor{catB}{B. Context aug.}} & \cellcolor{catC!15}\textbf{\textcolor{catC}{C. Composition}} & \cellcolor{catD!15}\textbf{\textcolor{catD}{D. Output proc.}} & \cellcolor{catE!15}\textbf{\textcolor{catE}{E. Compress.}} \\
\midrule

\cellcolor{catA!15}\textbf{\textcolor{catA}{A. Param.\ adapt.}}
& \cellcolor{compatOK!10}\textcolor{compatOK}{\textbf{Compatible}}: adapted model can also consume retrieved context
& \cellcolor{compatOK!10}\textcolor{compatOK}{\textbf{Compatible}}: pool members can be independently adapted (\S4.2, \S6.1)
& \cellcolor{cautionC!12}\textcolor{cautionC}{\textbf{Caution}}: TTA shifts error distribution, calibration set goes stale (\S4.3, \S7.2)
& \cellcolor{seqC!12}\textcolor{seqC}{\textbf{Sequential}}: distill/prune \emph{after} adaptation fixes the target model (\S8.3) \\

\cellcolor{catB!15}\textbf{\textcolor{catB}{B. Context aug.}}
& \cellcolor{gray!10}\centering---
& \cellcolor{compatOK!10}\textcolor{compatOK}{\textbf{Compatible}}: retrieval/memory can feed a routed model pool
& \cellcolor{cautionC!12}\textcolor{cautionC}{\textbf{Caution}}: retrieved context changes input distribution seen by calibrator
& \cellcolor{cautionC!12}\textcolor{cautionC}{\textbf{Caution}}: compression may reduce capacity to exploit long retrieved context \\

\cellcolor{catC!15}\textbf{\textcolor{catC}{C. Composition}}
& \cellcolor{gray!10}\centering---
& \cellcolor{gray!10}\centering---
& \cellcolor{cautionC!12}\textcolor{cautionC}{\textbf{Caution}}: per-model vs.\ per-ensemble calibration needs care
& \cellcolor{cautionC!12}\textcolor{cautionC}{\textbf{Caution}}: compressing individual experts trades pool diversity for efficiency \\

\cellcolor{catD!15}\textbf{\textcolor{catD}{D. Output proc.}}
& \cellcolor{gray!10}\centering---
& \cellcolor{gray!10}\centering---
& \cellcolor{gray!10}\centering---
& \cellcolor{cautionC!12}\textcolor{cautionC}{\textbf{Caution}}: distillation should preserve calibrated uncertainty, not just point forecasts (\S9.5) \\

\bottomrule
\end{tabular}
\caption{Pairwise interactions between the five post-training families.
\textcolor{compatOK}{\textbf{Compatible}} pairs compose without additional caveats;
\textcolor{cautionC}{\textbf{Caution}} pairs share an assumption that combining them can break;
\textcolor{seqC}{\textbf{Sequential}} indicates one is typically applied after the other by
design. Row/column colors follow the taxonomy families of Fig.~\ref{fig:tsfm_posttraining_taxonomy}.}
\label{tab:interactions}
\end{table}

\subsection{Data-Regime Sensitivity and Post-training Data Curation}
The taxonomy of Fig.~\ref{fig:tsfm_posttraining_taxonomy} organizes methods by locus of intervention, but several
families are implicitly sensitive to properties of the data regime itself --- series length,
sampling frequency, number of channels/variates, noise level, and the magnitude of distribution
shift between pretraining and target data. Table~\ref{tab:data-regime} summarizes this sensitivity
per family, drawing on evidence already presented in Sections~4--8.

Beyond method selection, the composition of the post-training set is itself a design choice that
recurs across families but is rarely made explicit. Three recurring considerations emerge from the
surveyed literature. First, \emph{diversity vs.\ specialization}: \work{MixFT}{mixft} partitions heterogeneous
fine-tuning data into more homogeneous sub-domains before training one LoRA module per sub-domain (\S4.2),
illustrating that an overly heterogeneous post-training set can dilute specialization, while an
overly narrow one risks the forgetting and overfitting failure modes discussed in \S4.4. Second,
\emph{sample filtering and confidence gating}: we identify
confidence-aware filtering of post-training samples as a general future direction for both retrieval and memory augmentation
(\S9.2). Third, \emph{domain matching}: federated post-training methods address cross-client domain
heterogeneity through aggregation strategies rather than pooling raw data (\S4.1), and
retrieval/memory methods construct domain- or task-specific knowledge bases rather than
general-purpose ones (\S5.1--5.2). We treat synthetic data augmentation for post-training as an
open direction rather than an established practice in the surveyed work: while synthetic corpora
are well studied for \emph{pretraining} TSFMs \citep{liu2025surveysynthetic}, their use for curating
\emph{post-training} sets --- e.g.\ to backfill underrepresented regimes identified via the
sensitivity axes above --- remains largely unexplored.

\begin{table}[t]
\centering
\small
\renewcommand{\arraystretch}{2.2}
\setlength{\tabcolsep}{5pt}
\begin{tabular}{
  >{\raggedright\arraybackslash}p{2.0cm}
  >{\raggedright\arraybackslash}p{2.5cm}
  >{\raggedright\arraybackslash}p{2.5cm}
  >{\raggedright\arraybackslash}p{2.5cm}
  >{\raggedright\arraybackslash}p{2.5cm}
  >{\raggedright\arraybackslash}p{2.5cm}
}
\toprule
\textbf{Family} & \textbf{Series length} & \textbf{Frequency} & \textbf{Channels} & \textbf{Noise} & \textbf{Shift magnitude} \\
\midrule

\rowcolor{catA!8}
\textcolor{catA}{\textbf{A. Param.\ adapt.}}
& \textcolor{lvlMed}{\textbf{Medium}}: multi-scale adapters target length (MSFT, \S4.2)
& \textcolor{lvlMed}{\textbf{Medium}}: cross-frequency handled explicitly in some PEFT
& \textcolor{lvlHigh}{\textbf{High}}: dedicated covariate/multivariate adapters (\S4.2)
& \textcolor{lvlHigh}{\textbf{High}}: forgetting risk under noisy target data (\S4.4)
& \textcolor{lvlHigh}{\textbf{High}}: full FT reserved for severe shift (\S4.4) \\

\rowcolor{catB!8}
\textcolor{catB}{\textbf{B. Context aug.}}
& \textcolor{lvlLow}{\textbf{Low}}--\textcolor{lvlMed}{\textbf{Med}}
& \textcolor{lvlLow}{\textbf{Low}}--\textcolor{lvlMed}{\textbf{Med}}
& \textcolor{lvlMed}{\textbf{Medium}}
& \textcolor{lvlHigh}{\textbf{High}}: irrelevant/noisy retrieval degrades predictions (\S5.4)
& \textcolor{lvlHigh}{\textbf{High}}: retrieval and transformation exist to bridge shift (\S5.3--\S5.4) \\

\rowcolor{catC!8}
\textcolor{catC}{\textbf{C. Composition}}
& \textcolor{lvlLow}{\textbf{Low}}
& \textcolor{lvlMed}{\textbf{Medium}}: pool diversity across domains/frequencies (\S6.1)
& \textcolor{lvlLow}{\textbf{Low}}--\textcolor{lvlMed}{\textbf{Med}}
& \textcolor{lvlLow}{\textbf{Low}}--\textcolor{lvlMed}{\textbf{Med}}
& \textcolor{lvlMed}{\textbf{Medium}}: routing implicitly assumes pool covers shift range \\

\rowcolor{catD!8}
\textcolor{catD}{\textbf{D. Output proc.}}
& \textcolor{lvlMed}{\textbf{Medium}}: multi-horizon calibration is harder over long horizons (\S7.2)
& \textcolor{lvlLow}{\textbf{Low}}
& \textcolor{lvlLow}{\textbf{Low}}--\textcolor{lvlMed}{\textbf{Med}}
& \textcolor{lvlMed}{\textbf{Medium}}
& \textcolor{lvlHigh}{\textbf{High}}: coverage guarantees weaken under shift (\S7.5) \\

\rowcolor{catE!8}
\textcolor{catE}{\textbf{E. Compression}}
& \textcolor{lvlMed}{\textbf{Medium}}: horizon-specific distillation objectives (\S8.1)
& \textcolor{lvlLow}{\textbf{Low}}--\textcolor{lvlMed}{\textbf{Med}}
& \textcolor{lvlLow}{\textbf{Low}}
& \textcolor{lvlLow}{\textbf{Low}}
& \textcolor{lvlMed}{\textbf{Medium}}: fixed pruned structure may not transfer across regimes (\S9.5) \\

\bottomrule
\end{tabular}
\caption{Qualitative sensitivity of each post-training family to data-regime properties,
grounded in evidence from Sections~4--8. Ratings follow a
\textcolor{lvlLow}{\textbf{Low}}~$\rightarrow$~\textcolor{lvlMed}{\textbf{Medium}}~$\rightarrow$~\textcolor{lvlHigh}{\textbf{High}}
sensitivity scale, colored for quick scanning; row labels follow the taxonomy families of
Fig.~\ref{fig:tsfm_posttraining_taxonomy}.}
\label{tab:data-regime}
\end{table}

\coloredsection{param!90!black}{Parameter Adaptation Methods}
\label{sec:parameter_adaptatation}

Parameter adaptation modifies the pretrained model parameters or attaches new trainable components to a frozen backbone. We organize the methods into three subcategories according to the {locus and timing of adaptation}:
full-parameter finetuning and continual pretraining, parameter-efficient finetuning, and test-time adaptation.
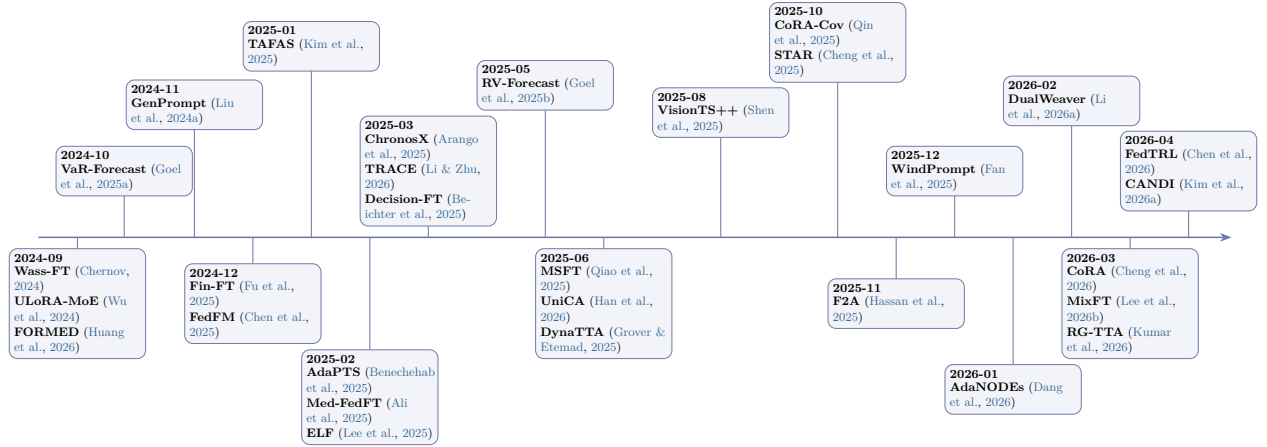
\begin{figure}[!h]
\centering
\resizebox{\textwidth}{!}{%
\begin{tikzpicture}[
    >=Latex,
    font=\footnotesize,
    timeline/.style={
        line width=1.15pt,
        draw=param!85!black
    },
    tick/.style={
        line width=0.75pt,
        draw=param!85!black
    },
    event/.style={
        rounded corners=6pt,
        align=left,
        text width=3.25cm,
        inner sep=3.5pt,
        font=\small,
        draw=param!85!black,
        fill=param!10
    },
    year/.style={
        font=\scriptsize\bfseries,
        text=param!85!black
    },
    labelbox/.style={
        rounded corners=5pt,
        align=center,
        font=\scriptsize\bfseries,
        text=white,
        fill=param!80!black,
        inner sep=4pt,
        minimum width=2.5cm
    }
]

\newcommand{\ABubble}[4]{%
    \draw[tick] (#1,0) -- (#1,#2);
    \node[event] at (#1,#2) {%
        \textbf{#3}\\[-1pt]
        #4
    };
}

% ...
\draw[timeline, -{Stealth[length=3mm]}] (-1.0,0) -- (29.6,0);




\ABubble{0}{-1.7}{2024-09}{
\textbf{Wass-FT}~\citep{wasserstein_ft}\\
\textbf{ULoRA-MoE}~\citep{uncertainty_ft}\\
\textbf{FORMED}~\citep{formed}
};

\ABubble{1.2}{1.7}{2024-10}{
\textbf{VaR-Forecast}~\citep{var_forecasting}\\
};

\ABubble{3.0}{3.4}{2024-11}{
\textbf{GenPrompt}~\citep{generalized_prompt_tuning}
};

\ABubble{4.5}{-1.7}{2024-12}{
\textbf{Fin-FT}~\citep{financial_finetuning}
};

\ABubble{6.0}{4.9}{2025-01}{
\textbf{TAFAS}~\citep{tafas}
};

\ABubble{7.5}{-4.1}{2025-02}{
\textbf{AdaPTS}~\citep{adapts}\\
\textbf{Med-FedFT}~\citep{medical_federated_finetuning}\\
\textbf{ELF}~\citep{elf}
};

\ABubble{9.0}{1.7}{2025-03}{
\textbf{ChronosX}~\citep{chronosx}\\
\textbf{TRACE}~\citep{trace}\\
\textbf{Decision-FT}~\citep{decision_focused_ft}
};

\ABubble{12.0}{3.9}{2025-05}{
\textbf{RV-Forecast}~\citep{realized_volatility}
};

\ABubble{13.5}{-1.7}{2025-06}{
\textbf{MSFT}~\citep{msft}\\
\textbf{UniCA}~\citep{unica}
};

\ABubble{16.5}{3.2}{2025-08}{
\textbf{VisionTS++}~\citep{visiontspp}
};


\ABubble{19.5}{5}{2025-10}{
\textbf{CoRA-Cov}~\citep{cora_covariate}\\
\textbf{STAR}~\citep{star}
};

\ABubble{21.0}{-1.7}{2025-11}{
\textbf{F2A}~\citep{f2a}
};

\ABubble{22.5}{1.7}{2025-12}{
\textbf{WindPrompt}~\citep{wind_power_prompt}
};

\ABubble{25.5}{3.5}{2026-02}{
\textbf{DualWeaver}~\citep{dualweaver}
};

\ABubble{27.0}{-1.7}{2026-03}{
\textbf{CoRA}~\citep{cora}\\
\textbf{MixFT}~\citep{mixft}
};

\end{tikzpicture}%
}
\caption{Timeline of parameter adaptation methods for post-training of TSFMs.}
\label{fig:parameter-adaptation-timeline}
\end{figure}

\subsection{Full-parameter finetuning and Continual Pretraining}
Full-parameter finetuning and continual pretraining adapt a pretrained TSFM by updating a large portion of its parameters. Full-parameter finetuning typically uses supervised target-domain objectives, whereas continual pretraining continues self-supervised or forecasting-oriented training on additional temporal data. Both approaches are more expensive than lightweight adaptation methods, but offer greater capacity for domain specialization. Methods in this category differ primarily in the adaptation objective, the data used for updating the model, and whether the goal is task-specific specialization or broader domain adaptation.

\textcolor{black}{Some approaches tailor the full-parameter adaptation objective to the target task. Wasserstein finetuning replaces the standard cross-entropy objective with a distance-aware loss that improves point-estimation accuracy, although it may degrade probabilistic forecasting quality as measured by weighted quantile loss~\citep{wasserstein_ft}.}

Full-parameter finetuning is also widely used for domain-specific adaptation. In finance, large time series models are continually pretrained for market forecasting, where the target distribution is highly nonstationary and can differ substantially from general-purpose pretraining data~\citep{financial_finetuning}. When target-domain data cannot be centralized, federated post-training adapts TSFMs across decentralized clients without sharing raw data. Federated medical time-series studies combine local updates with global aggregation to address privacy constraints and client heterogeneity~\citep{medical_federated_finetuning}. Cross-modal continual pretraining methods, such as VisionTS++~\citep{visionts,visiontspp}, further show that backbones pretrained in other modalities can be adapted to time series by continuing training on large-scale temporal data; we include it as a boundary case because its ImageNet-pretrained starting point is already used zero-shot as a forecaster.

Overall, full-parameter finetuning and continual pretraining offer a high-capacity
form of TSFM adaptation among the methods surveyed, allowing pretrained models to be reshaped for
new domains, tasks, and data regimes. This flexibility is particularly valuable when the target distribution differs substantially from the pretraining data or when the adaptation objective departs from generic forecasting accuracy. At the same time, these methods are computationally expensive, data-dependent, and vulnerable to overfitting, catastrophic forgetting, and proliferation of domain-specific model copies. These trade-offs motivate parameter-efficient and modular post-training strategies that seek similar specialization with lower adaptation and deployment cost.

\subsection{Parameter-Efficient Adaptation}

Parameter-efficient adaptation specializes a pretrained TSFM while keeping most or all of its original parameters frozen. These methods optimize a small subset of existing parameters or introduce lightweight trainable components, reducing adaptation cost and allowing multiple target settings to share the same pretrained backbone.

A first branch makes PEFT scale- or objective-aware while keeping the pretrained parameters frozen. MSFT~\citep{msft} constructs multi-scale inputs and trains scale-specific adapters, LoRA modules, cross-scale aggregators, and output-mixing weights to exploit temporal patterns at different resolutions. Decision-focused finetuning~\citep{ decision_focused_ft} instead trains LoRA or DoRA adapters on Moirai using a differentiable surrogate of downstream dispatch costs, thereby optimizing decision quality rather than conventional forecasting errors.

A second branch extends the variable interface of pretrained TSFMs. For covariate-aware forecasting, ChronosX~\citep{chronosx}, UniCA~\citep{unica}, and covariate-aware CoRA~\citep{cora_covariate} incorporate past, future, or heterogeneous covariates through modular
injection and fusion mechanisms. Application-specific prompt tuning similarly encodes exogenous meteorological variables as time-series prompts for wind-power forecasting~\citep{wind_power_prompt}.
For multivariate adaptation, correlation-aware CoRA~\citep{cora}
models time-varying and time-invariant inter-channel dependencies, while AdaPTS~\citep{adapts} projects multivariate inputs into TSFM-compatible latent series. DualWeaver~\citep{dualweaver} constructs learnable surrogate series through cross-variable feature fusion, and
generalized prompt tuning~\citep{generalized_prompt_tuning} uses trainable prompts to combine information across channels while keeping the univariate backbone frozen.

A third branch addresses data heterogeneity and task specialization. MixFT~\citep{mixft} partitions heterogeneous time series into more homogeneous groups and adapts lightweight modules to each group. TRACE~\citep{trace} studies PEFT strategies across forecasting and anomaly detection, while STAR~\citep{star} introduces state-aware adapters for multivariate anomaly detection. For anomaly-related tasks, Forecast2Anomaly~\citep{f2a} finetunes only the decoder and task heads of a TSFM with a frozen encoder for anomaly prediction, and ULoRA-MoE~\citep{uncertainty_ft} trains a mixture of LoRA experts with an uncertainty-aware Gumbel-Softmax router. FORMED~\citep{formed}
repurposes a frozen TSFM for medical time-series classification through lightweight task-specific adaptation.
In finance, VaR-Forecast~\citep{var_forecasting} and RV-Forecast~\citep{realized_volatility} adapt TimesFM to value-at-risk and realized-volatility forecasting by linear probing, keeping the transformer layers frozen.

Overall, parameter-efficient adaptation extends pretrained TSFMs to new temporal scales, objectives, variables, domains, and tasks without retraining the full backbone. Its effectiveness nevertheless depends on whether the frozen model contains reusable temporal knowledge and whether the added modules, their placement, and their optimization objectives are sufficient for the target setting \citep{xiao2026instanceadaptivepromptscontexttimeseries}.

\subsection{Test-Time and Online Adaptation}

Test-Time and Online Adaptation updates a model component during deployment, allowing the model to respond to nonstationarity, regime shifts, and streaming feedback. ELF trains lightweight forecasters and weighting modules online while keeping the TSFM backbone fixed~\citep{elf}.
{Some methods focus on lightweight online updating.}
TAFAS studies nonstationary forecasting with test-time adaptation, training input and output gated calibration modules around a frozen forecaster on partially observed ground truth; it is evaluated mainly with per-dataset forecasters and, in one experiment, with Chronos~\citep{tafas}. 
Related test-time adaptation methods, namely AdaNODEs~\citep{adanodes}, which models continuous drift with Neural ODEs, and DynaTTA~\citep{dynatta} and RG-TTA~\citep{rgtta}, which use shift- or regime-guided controllers to decide when and how strongly to adapt, were developed for task-specific forecasters and have not been evaluated on TSFMs; we mention them as transferable techniques but exclude them from our catalogue.

Taken together, these methods highlight the central trade-off in test-time and online adaptation: TSFMs must remain flexible enough to track nonstationarity, while avoiding unstable updates caused by noise, transient fluctuations, or anomalous observations. Recent work therefore moves beyond unconstrained online updating toward controlled adaptation, using lightweight modules or calibration gates, with the drift models and regime detectors of the related methods above as candidate extensions for TSFMs. This makes test-time adaptation a promising but delicate post-training strategy, whose success depends on balancing responsiveness to target-domain change with preservation of the pretrained model's general knowledge.

\subsection{Development and Limitations}
Parameter adaptation methods have evolved from direct finetuning toward increasingly lightweight and deployment-aware strategies as shown in~\cref{fig:parameter-adaptation-timeline}. Full-parameter finetuning and continual pretraining offer strong domain specialization but demand substantial computation and target-domain data. Parameter-efficient methods reduce this cost by optimizing only a small number of parameters on top of frozen backbones. 
Test-time and online adaptation further shift the adaptation process into deployment, trading offline training cost for the ability to respond to nonstationary streams.

Despite their effectiveness, parameter adaptation methods also face several limitations. Full-parameter finetuning can be computationally expensive and may cause catastrophic forgetting, particularly when the target domain is scarce or noisy. Parameter-efficient methods reduce cost but introduce architectural decisions, such as where to insert adapters, how to encode
covariates, and how to model cross-channel dependence, which remain largely task-specific and lack principled selection criteria. 
Test-time adaptation can be unstable when the test stream contains anomalies, delayed labels, abrupt regime changes, or weak self-supervised signals.
{Changing the adaptation objective can also trade one forecasting property for another: Wasserstein finetuning improves point accuracy but can degrade weighted quantile loss~\citep{wasserstein_ft}, so gains reported under a single metric do not show that the pretrained model's probabilistic forecasts are preserved.
Table~\ref{tab:data-regime} summarizes qualitatively when each subfamily is indicated: full
fine-tuning is indicated only under high shift magnitude and sufficient data volume, while PEFT
and test-time adaptation are preferred as noise and shift sensitivity increase relative to
available target data.}

\subsection{Domain- and Frequency-Transfer Mechanisms}
Several parameter-adaptation methods target domain or sampling-frequency mismatch specifically,
rather than adaptation in general. AdaPTS projects multivariate inputs from a new domain into a
TSFM-compatible latent space (\S4.2), and covariate-aware CoRA (CoRA-Cov) models domain-specific
covariate structure (\S4.2). More broadly, early frequency-specific transfer-learning work showed
that models trained at one sampling frequency do not generalize directly to another without
target-side adaptation (\S1), motivating adapter designs that explicitly encode frequency as an
adaptation axis rather than treating it as an incidental input property. Unifying these
frequency- and domain-transfer mechanisms into a shared design space is an open problem: no
current method reports transfer performance as a function of both shift type simultaneously.

\subsection{Federated and Privacy-Preserving Post-training}
When target-domain data cannot be centralized --- common in healthcare and finance --- federated
post-training adapts a TSFM across decentralized clients without sharing raw data. Med-FedFT
applies this to medical time series by finetuning a pretrained Chronos model (\S4.1). This work
builds on finetuning the pretrained model; extending
federated post-training to PEFT, retrieval, or output calibration remains unexplored, and
the surveyed work does not report formal privacy guarantees (e.g. differential privacy) alongside
forecasting accuracy.

\coloredsection{context!90!black}{Context Augmentation Methods}
\label{sec:context_augmentation}

Context augmentation constitutes an input-level post-training paradigm for TSFMs, which improves TSFMs by providing additional information alongside the target input, without modifying model parameters. The central question is how to construct useful contexts for each target input.
{We identify three subcategories by the structure of the context: retrieval augmentation, memory augmentation, and context transformation.}
\begin{figure}[!h]
\centering
\resizebox{\textwidth}{!}{%
\begin{tikzpicture}[
    >=Latex,
    font=\footnotesize,
    timeline/.style={
        line width=1.1pt,
        draw=context!85!black
    },
    tick/.style={
        line width=0.8pt,
        draw=context!85!black
    },
    bevent/.style={
        rounded corners=6pt,
        align=left,
        text width=2.45cm,
        inner sep=3pt,
        font=\scriptsize,
        draw=context!85!black,
        fill=context!12
    },
    year/.style={
        font=\scriptsize\bfseries,
        text=context!85!black
    },
    legend/.style={
        rounded corners=4pt,
        draw=context!85!black,
        fill=context!12,
        inner sep=2.5pt,
        font=\scriptsize,
        align=center
    }
]


\draw[timeline, -{Stealth[length=3mm]}] (-0.8,0) -- (20.6,0);


\foreach \x in {0,2.2,4.4,6.6,8.8,11.0,13.2,15.4,17.6,19.8} {
    \draw[tick] (\x,0) -- (\x,-0);
}





\draw[tick] (0,0) -- (0,0.48);
\node[bevent, anchor=south] at (0,0.55) {
\textbf{2024-11}\\[-1pt]
\textbf{RAF}~\citep{raf}
};

\draw[tick] (2.2,0) -- (2.2,-0.48);
\node[bevent, anchor=north] at (2.2,-0.55) {
\textbf{2024-12}\\[-1pt]
\textbf{TimeRAF}~\citep{timeraf}
};

\draw[tick] (4.4,0) -- (4.4,0.48);
\node[bevent, anchor=south] at (4.4,0.55) {
\textbf{2025-03}\\[-1pt]
\textbf{TS-RAG}~\citep{tsrag}
};

\draw[tick] (6.6,0) -- (6.6,-0.48);
\node[bevent, anchor=north] at (6.6,-0.55) {
\textbf{2025-06}\\[-1pt]
\textbf{RATFM}~\citep{ratfm}
};

\draw[tick] (8.8,0) -- (8.8,0.48);
\node[bevent, anchor=south] at (8.8,0.55) {
\textbf{2025-09}\\[-1pt]
\textbf{MOMEMTO} \citep{momemto}
};

\draw[tick] (11.0,0) -- (11.0,-0.48);
\node[bevent, anchor=north] at (11.0,-0.55) {
\textbf{2026-02}\\[-1pt]
\textbf{TS-Memory}~\citep{tsmemory}
};

\draw[tick] (13.2,0) -- (13.2,0.48);
\node[bevent, anchor=south] at (13.2,0.55) {
\textbf{2026-02}\\[-1pt]
\textbf{MEMTS}~\citep{memts}
};

\draw[tick] (15.4,0) -- (15.4,-0.48);
\node[bevent, anchor=north] at (15.4,-0.55) {
\textbf{2026-02}\\[-1pt]
\textbf{TATO}~\citep{tato}
};

\draw[tick] (17.6,0) -- (17.6,0.48);
\node[bevent, anchor=south] at (17.6,0.55) {
\textbf{2026-03}\\[-1pt]
\textbf{RAG4CTS}~\citep{RAG4CTS}
};

\draw[tick] (19.8,0) -- (19.8,-0.48);
\node[bevent, anchor=north] at (19.8,-0.55) {
\textbf{2026-03}\\[-1pt]
\textbf{Cross-RAG}~\citep{crossrag}
};

\end{tikzpicture}%
}
\caption{Timeline of context augmentation methods for post-training of TSFMs.}
\label{fig:context-augmentation-timeline}
\end{figure}
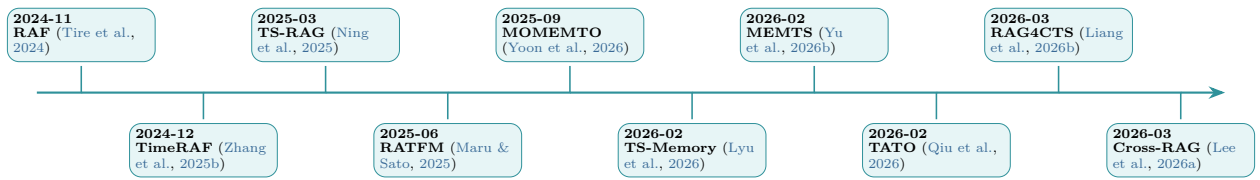

\subsection{Retrieval Augmentation}
Retrieval augmentation methods retrieve related historical series or segments and inject them into the forecasting or detection process. 
\textcolor{black}{Retrieval Augmented Forecasting (RAF; \cite{raf}) augments zero-shot TSFM forecasts by retrieving related time series examples and incorporating them into the forecasting process. The reported gains are observed across diverse domains.}
TimeRAF~\citep{timeraf} further develops this idea by constructing customized time series knowledge bases for specific forecasting tasks. It uses a learnable retriever to extract useful information and introduces channel prompting to integrate the retrieved knowledge along the channel dimension.
TS-RAG~\citep{tsrag} also retrieves semantically relevant time series segments from a dedicated knowledge database, but focuses on fusing the retrieved patterns with the TSFM representation through a learnable Adaptive Retrieval Mixer. Cross-RAG~\citep{crossrag} observes that a fixed set of retrieved samples may contain irrelevant information. It therefore introduces query--retrieval cross attention to model input-level relevance between the target series and the retrieved examples. This allows the model to selectively attend to useful retrieved samples. RATFM~\citep{ratfm} extends retrieval augmentation from forecasting to anomaly detection. It retrieves normal examples from the target domain and uses them as test-time adaptation signals, enabling pretrained TSFMs to approach in-domain performance without domain-dependent parameter updates.
RAG4CTS~\citep{RAG4CTS} studies retrieval augmentation for covariate time series in industrial predictive maintenance. 

Taken together, these methods show that retrieval augmentation provides a non-parametric route to TSFM post-training: instead of updating the backbone, the model is supplied with relevant historical examples, normal patterns, or task-specific temporal knowledge at inference time. The central design questions are what to retrieve, how to define relevance, and how to fuse retrieved information with the target series. Recent work therefore moves from simple retrieval of related examples toward learned retrievers, customized knowledge bases, cross-attention relevance modeling, and mixture-of-experts fusion. This makes retrieval augmentation especially attractive when target-domain data are available but parameter updates are costly or risky. However, the approach remains sensitive to retrieval quality, since irrelevant or distributionally mismatched examples may introduce noise; robust retrieval, filtering, and selective fusion are therefore key open challenges.

\subsection{Memory Augmentation}
Memory augmentation replaces explicit retrieval with memory modules that store domain knowledge or normal patterns in a compact form. MEMTS~\citep{memts} addresses the latency and scalability limitations of retrieval-augmented adaptation. 
The method introduces a Knowledge Persistence Module that internalizes domain-specific temporal dynamics, such as recurring seasonal patterns and
trends, into learnable latent prototypes. 
TS-Memory~\citep{tsmemory} first builds a kNN teacher that produces confidence-aware quantile targets from the retrieved futures, then distills the retrieval correction into a lightweight memory adapter. In inference, the memory adapter fuses with the backbone prediction.
Memory augmentation has also been applied to anomaly detection. MOMEMTO~\citep{momemto} introduces a patch-based memory module that stores representative normal patterns to solve the problem that high-capacity reconstruction models may over-generalize and reconstruct unseen anomalies too well. The memory gate constrains reconstruction through normal prototypes and improves anomaly detection, especially in few-shot and multi-domain settings.

Overall, memory augmentation provides a scalable alternative to retrieval-based post-training by compressing useful temporal knowledge into learned prototypes, adapters, or memory banks. This makes memory augmentation attractive when retrieval is too costly or when compact domain specialization is needed. Its effectiveness, however, depends on how the memory is constructed and maintained: prototypes must capture stable target-domain structure while avoiding noise, contamination, and overfitting to narrow regimes. Robust memory update rules, uncertainty-aware prototype selection, and mechanisms for forgetting outdated patterns remain important open directions.

\subsection{Context Transformation}
Context transformation adapts a frozen TSFM by transforming the target input before prediction. Given an input series \(\mathbf{x}_{1:T}\), these methods learn or search for a transformation \(\tau_\eta\) such that
\[
    \mathbf{z}_{1:T'}
    =
    \tau_\eta(\mathbf{x}_{1:T}),
    \qquad
    \hat{\mathbf{y}}_{1:H}
    =
    f_{\theta_0}(\mathbf{z}_{1:T'}).
\]
The transformation may change how the context is sliced, normalized, denoised, corrected, or represented. TATO~\citep{tato} is a representative example. Related input-side methods, such as the input nudging component of \(\delta\)-Adapter~\citep{delta_adapter}, also suggest that small transformations at the model interface can improve forecasting without modifying the deployed backbone.

Context transformation is most useful when the pretrained backbone must remain fixed, but the target input can be adapted to better match the model interface. The main open challenge is to design transformations that correct distributional mismatch while preserving task-relevant temporal information.

\subsection{Development and Limitations}

Context augmentation methods have evolved from injecting external evidence into the prediction process to reshaping the input representation itself, as summarized in~\cref{fig:context-augmentation-timeline}. Retrieval-augmented methods retrieve related historical examples from external knowledge bases and incorporate them into forecasting or detection. Memory augmentation compresses such external knowledge into compact modules, reducing the need for explicit retrieval at inference time. Context transformation goes one step further by adapting the input representation directly, without adding external examples or modifying the pretrained backbone.

Despite this progression, context augmentation methods face several limitations. First, their effectiveness depends strongly on the quality of the augmented context. Irrelevant retrieved examples can introduce noise and degrade predictions, especially in sparse or poorly matched target domains. Second, retrieval-based methods introduce memory, indexing, and latency overheads that may be prohibitive in real-time settings. Memory-based methods mitigate these costs, but may over-compress domain knowledge or struggle to update efficiently under distribution shift. Third, retrieval is least informative under the very shift it is meant to address: because the knowledge base is built from past data, a test window from a regime not yet represented in it is unlikely to have a close neighbour. Finally, when retrieved segments are concatenated with the input, as in RAF~\citep{raf}, they compete with the target history for the backbone's bounded context length. 

\coloredsection{composition!90!black}{Model Composition Methods}
\label{sec:model_composition}

Model composition methods constitute a model-level post-training
paradigm for TSFMs. 
This line of work maintains a pool of pretrained, post-trained, or statistically enhanced forecasters and learns how to allocate predictive responsibility among them for a given target time series. The central question is determining which models to select and how to weight their contributions.
{We identify three subcategories by how the allocation is performed: model selection, adaptive fusion, and sequential fusion.}
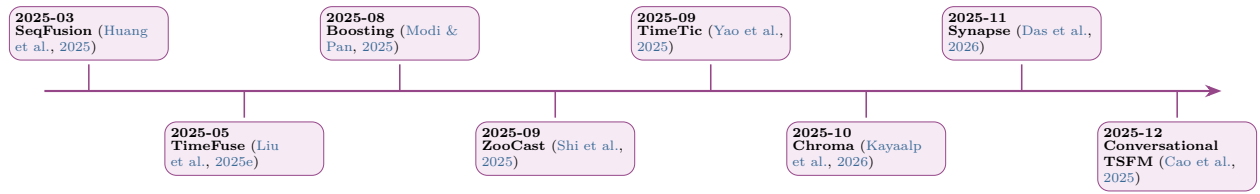
\begin{figure}[!h]
\centering
\resizebox{\textwidth}{!}{%
\begin{tikzpicture}[
    >=Latex,
    font=\footnotesize,
    timeline/.style={
        line width=1.1pt,
        draw=composition!85!black
    },
    tick/.style={
        line width=0.8pt,
        draw=composition!85!black
    },
    cevent/.style={
        rounded corners=6pt,
        align=left,
        text width=2.65cm,
        inner sep=3pt,
        font=\scriptsize,
        draw=composition!85!black,
        fill=composition!12
    },
    year/.style={
        font=\scriptsize\bfseries,
        text=composition!85!black
    },
    legend/.style={
        rounded corners=4pt,
        draw=composition!85!black,
        fill=composition!12,
        inner sep=2.5pt,
        font=\scriptsize,
        align=center
    }
]


\draw[timeline, -{Stealth[length=3mm]}] (-0.8,0) -- (20.4,0);


\foreach \x in {0,2.8,5.6,8.4,11.2,14.0,16.8,19.6} {
    \draw[tick] (\x,0) -- (\x,-0);
}


\draw[tick] (0,0) -- (0,0.48);
\node[cevent, anchor=south] at (0,0.55) {
\textbf{2025-03}\\[-1pt]
\textbf{SeqFusion}~\citep{seqfusion}
};

\draw[tick] (5.6,0) -- (5.6,0.48);
\node[cevent, anchor=south] at (5.6,0.55) {
\textbf{2025-08}\\[-1pt]
\textbf{Boosting}~\citep{boosting}
};

\draw[tick] (8.4,0) -- (8.4,-0.48);
\node[cevent, anchor=north] at (8.4,-0.55) {
\textbf{2025-09}\\[-1pt]
\textbf{ZooCast}~\citep{zoocast}
};

\draw[tick] (11.2,0) -- (11.2,0.48);
\node[cevent, anchor=south] at (11.2,0.55) {
\textbf{2025-09}\\[-1pt]
\textbf{TimeTic}~\citep{timetic}
};

\draw[tick] (14.0,0) -- (14.0,-0.48);
\node[cevent, anchor=north] at (14.0,-0.55) {
\textbf{2025-10}\\[-1pt]
\textbf{Chroma}~\citep{chroma}
};

\draw[tick] (16.8,0) -- (16.8,0.48);
\node[cevent, anchor=south] at (16.8,0.55) {
\textbf{2025-11}\\[-1pt]
\textbf{Synapse}~\citep{synapse}
};

\draw[tick] (19.6,0) -- (19.6,-0.48);
\node[cevent, anchor=north] at (19.6,-0.55) {
\textbf{2025-12}\\[-1pt]
\textbf{Conversational TSFM}~\citep{conversational}
};

\end{tikzpicture}%
}
\caption{Timeline of model composition methods for post-training of TSFMs.}
\label{fig:model-composition-timeline}
\end{figure}
\subsection{Model Selection}
Model selection methods select one or a small subset of models from a candidate pool for a given target series, dataset, or task.
Chroma \citep{chroma} investigates whether a set of smaller pretrained forecasting models can replace a single large TSFM. 
It shows that ensembling or model selection over a model bank, which consists of diverse specialists post-trained from a shared base model, can achieve competitive forecasting performance with substantially fewer activated parameters. ZooCast~\citep{zoocast} embeds models in a unified representation space and selects suitable forecasters through similarity matching.
This design is attractive when new TSFMs are continuously released, since the model zoo can be expanded without retraining the entire selection system.
TimeTic~\citep{timetic} extends model selection from zero-shot performance prediction to transferability estimation. Given observed model--dataset--performance relationships on source datasets, it predicts how a TSFM will perform after finetuning on an unseen target dataset, selecting the model expected to yield the strongest downstream adaptation.

Model selection is most useful when the candidate pool contains genuinely complementary models, but its reliability depends on estimating model suitability from limited target-domain evidence. The main open challenge is to develop selection criteria that remain robust under distribution shift and evolving model zoos.

\subsection{Adaptive Fusion}
Adaptive fusion methods combine the predictions of multiple models using input-dependent or context-dependent weights. TimeFuse~\citep{timefuse}, which extracts meta-features from the input series and trains a fusor to predict sample-level fusion weights, was developed for task-specific forecasters trained on each dataset rather than for TSFMs; we mention it as a transferable technique but exclude it from our catalogue. 
Synapse \citep{synapse} focuses on the complementary expertise of TSFMs. It dynamically assigns predictive weights according to context performance and constructs a robust forecast distribution by adaptively sampling from the output quantiles of constituent models. 
\citet{boosting} revisit classical ensemble and statistical methods for TSFM, improving forecasting robustness through bagging, stacking, residual modeling, prediction intervals, and iterative error feedback.

Unlike model selection, which commits to one or a few forecasters, adaptive fusion keeps multiple models active and combines them through input- or context-dependent weights. This can better exploit complementary model behavior and provide smoother adaptation across regimes, but it also increases inference cost and may become unstable when the weighting signal is noisy or distributionally shifted. A central open question is how to make this additional flexibility translate into consistent gains over model selection while keeping inference costs under control.

\subsection{Sequential Fusion}
Sequential fusion treats model composition as a multi-step decision process.
SeqFusion \citep{seqfusion} selects suitable pretrained models according to the temporal characteristics of the target series, invokes them sequentially, and fuses their predictions; its zoo consists mainly of small single-dataset models, with a TSFM zoo (Chronos, Moirai, TimesFM) evaluated in one experiment. 
Conversational TSFM \citep{conversational} pushes this direction further toward agentic orchestration. It positions an LLM as a judge that evaluates, explains, and coordinates an ensemble of TSFMs. The LLM is trained to associate ensemble weights with interpretable temporal reasoning and then refines its decisions through multi-turn interaction. 

Sequential fusion extends model composition from one-shot routing to staged orchestration, allowing later decisions to depend on intermediate forecasts, diagnostics, or model explanations. This can improve flexibility and interpretability, but it also introduces additional latency, error propagation, and dependence on the reliability of the controller. A key open question is when multi-step orchestration provides measurable gains over simpler selection or fusion strategies, especially under realistic inference-cost constraints.

\subsection{Development and Limitations}
Model Composition has evolved from a simple combination to structured orchestration, as shown in~\cref{fig:model-composition-timeline}. Early approaches apply classical strategies such as averaging and bagging. Subsequent methods introduce input-conditional fusion, sequential selection pipelines, and transferability-based model selection. Recent agentic methods further add iterative reasoning, enabling an LLM to select, explain, and refine model combinations through multi-turn interaction.

Despite the advances, model composition remains at an early stage. First, many routers rely on shallow signals, such as handcrafted meta features, representation distances, historical validation performance, or learned ensemble weights, which may not capture causal dynamics and regime changes. Second, the inference cost of composition grows with the number of models executed per query, from a single model under selection up to the whole pool under fusion, yet activated parameters, memory, and latency are not reported consistently, so accuracy gains cannot be weighed against this cost across methods. 
Third, interpretability remains limited. Although agentic methods attempt to explain routing decisions, most current methods do not verify whether the learned weights reflect meaningful temporal properties rather
than spurious benchmark correlations. Finally, the current literature is dominated by forecasting, while routing across anomaly detection, classification, and imputation tasks remains underexplored.

\coloredsection{output!90!black}{Output Processing and Uncertainty Control Methods}
\label{sec:output_processing_uncertainty}

Output processing provides an output-level paradigm for TSFM post-training. Rather than modifying the pretrained backbone or its input context, these methods operate after a model has produced an initial forecast, predictive distribution, reconstruction, or anomaly score. The central question is how to improve the accuracy, calibration, reliability, or decision utility of these outputs through post-hoc refinement. We organize this category into four subfamilies according to the object being refined and the strategy used: forecast refinement, forecast calibration, probabilistic output modeling, and anomaly detection.
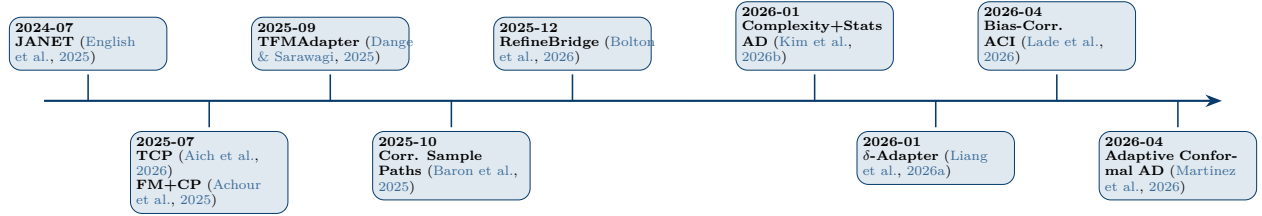
\begin{figure}[!h]
\centering
\resizebox{\textwidth}{!}{%
\begin{tikzpicture}[
    >=Latex,
    font=\footnotesize,
    timeline/.style={
        line width=1.1pt,
        draw=output!85!black
    },
    tick/.style={
        line width=0.8pt,
        draw=output!85!black
    },
    devent/.style={
        rounded corners=6pt,
        align=left,
        text width=2.65cm,
        inner sep=3pt,
        font=\scriptsize,
        draw=output!85!black,
        fill=output!12
    },
    year/.style={
        font=\scriptsize\bfseries,
        text=output!85!black
    },
    legend/.style={
        rounded corners=4pt,
        draw=output!85!black,
        fill=output!12,
        inner sep=2.5pt,
        font=\scriptsize,
        align=center
    }
]


\draw[timeline, -{Stealth[length=3mm]}] (-0.8,0) -- (20.6,0);


\foreach \x in {0,2.2,4.4,6.6,8.8,11.0,13.2,15.4,17.6,19.8} {
    \draw[tick] (\x,0) -- (\x,-0);
}


\draw[tick] (2.2,0) -- (2.2,-0.48);
\node[devent, anchor=north] at (2.2,-0.55) {
\textbf{2025-07}\\[-1pt]
\textbf{FM+CP}~\citep{fm_conformal_prediction}\\[-1pt]
};

\draw[tick] (4.4,0) -- (4.4,0.48);
\node[devent, anchor=south] at (4.4,0.55) {
\textbf{2025-09}\\[-1pt]
\textbf{TFMAdapter}~\citep{tfmadapter}\\[-1pt]
};


\draw[tick] (6.6,0) -- (6.6,-0.48);
\node[devent, anchor=north] at (6.6,-0.55) {
\textbf{2025-10}\\[-1pt]
\textbf{Corr. Sample Paths}~\citep{correlated_sample_paths}\\[-1pt]
};

\draw[tick] (8.8,0) -- (8.8,0.48);
\node[devent, anchor=south] at (8.8,0.55) {
\textbf{2025-12}\\[-1pt]
\textbf{RefineBridge}~\citep{refinebridge}\\[-1pt]
};


\draw[tick] (13.2,0) -- (13.2,0.48);
\node[devent, anchor=south] at (13.2,0.55) {
\textbf{2026-01}\\[-1pt]
\textbf{Complexity+Stats AD}~\citep{complexity_statistics}\\[-1pt]
};

\draw[tick] (15.4,0) -- (15.4,-0.48);
\node[devent, anchor=north] at (15.4,-0.55) {
\textbf{2026-01}\\[-1pt]
\textbf{\(\delta\)-Adapter}~\citep{delta_adapter}\\[-1pt]
};

\draw[tick] (19.8,0) -- (19.8,-0.48);
\node[devent, anchor=north] at (19.8,-0.55) {
\textbf{2026-04}\\[-1pt]
\textbf{Adaptive Conformal AD}~\citep{adaptive_conformal}\\[-1pt]
};


\end{tikzpicture}%
}
\caption{Timeline of output processing and uncertainty control methods for post-training of TSFMs.}
\label{fig:output-uncertainty-timeline}
\end{figure}

\subsection{Forecast Refinement}

Forecast refinement methods directly adjust the initial forecast produced by a TSFM. Given an initial prediction
\(
    \tilde{\mathbf{y}}_{1:H}
    =
    f_{\theta_0}(\mathbf{x}_{1:T}),
\)
these methods learn a post-processing map
\[
    \hat{\mathbf{y}}_{1:H}
    =
    \tilde{\mathbf{y}}_{1:H}
    +
    g_\psi
    \left(
    \mathbf{x}_{1:T},
    \tilde{\mathbf{y}}_{1:H}
    \right),
\]
where \(g_\psi\) estimates a residual correction that compensates for systematic errors, biases, or distributional mismatch in the raw forecast. The Forecast After the Forecast~\citep{delta_adapter} proposes \(\delta\)-Adapter, a lightweight post-processing framework that combines input nudging, output residual correction, and uncertainty calibration. RefineBridge~\citep{refinebridge} treats TSFM forecasts as generative priors and uses a context-conditioned Schr\"odinger Bridge refinement module to transport the initial forecast toward the target distribution. \textcolor{black}{TFMAdapter~\citep{tfmadapter} incorporates future covariates without finetuning the TSFM. It combines univariate TSFM forecasts, regression-based pseudo-forecasts, and covariates through an instance-level non-parametric refinement cascade; because this adapter is refitted on each query's history, we include it as a boundary case of our persistence criterion (\S2).}

Forecast refinement is most valuable when TSFM errors contain stable, target-specific structure such as horizon-dependent bias, scale mismatch, or recurring residual patterns. The main challenge is to estimate the residual map \(g_\psi\) from limited target data while detecting cases where the raw TSFM forecast should be left unchanged.

\subsection{Forecast Calibration}
Forecast calibration methods transform TSFM forecasts into calibrated prediction intervals or joint prediction regions. The primary goal is to provide uncertainty sets with desired coverage properties. Foundation models for time series forecasting with conformal prediction
\citep{fm_conformal_prediction} investigates how zero-shot TSFMs can benefit conformal prediction, since more target domain samples can be reserved for calibration rather than model training. 
Related conformal methods, namely Temporal Conformal Prediction~\citep{tcp}, which combines rolling split-conformal calibration with a Robbins--Monro update, bias-corrected adaptive conformal inference~\citep{Bias-Corrected}, which re-centers intervals by estimating online forecast bias, and JANET~\citep{janet}, which constructs joint prediction regions across horizons, were developed for non-TSFM forecasters (gradient-boosted trees, ridge/ARIMA, and AR/LSTM models, respectively) and have not been evaluated on TSFMs; we mention them as transferable techniques but exclude them from our catalogue.

In this setting, calibration acts as a lightweight reliability layer on top of TSFM forecasts, correcting coverage without retraining the backbone. The main challenge is to maintain coverage under temporal dependence, nonstationarity, and multi-horizon error correlation while keeping prediction sets sharp. This makes calibration-data efficiency, online bias correction, and joint-region construction central directions for TSFM uncertainty control.

\subsection{Probabilistic Output Modeling}
Probabilistic output modeling methods construct, parameterize or refine the predictive distribution of a TSFM. Unlike conformal calibration that focuses on coverage guarantees, these methods aim to model distributional structure, uncertainty decomposition, or temporal dependence among future horizons.
Efficiently Generating Correlated Sample Paths~\citep{correlated_sample_paths} addresses a different limitation of multi-step TSFMs: many models provide marginal distributions for each future horizon but ignore the joint dependence across horizons. The method introduces a copula-based post-processing layer to generate correlated sample paths from existing multi-step TSFM outputs in a single forward pass; its main variant sets the copula correlation from each series' context autocorrelation, while a trained variant learns it across datasets, which is the persistent component that places the method within our scope (\S2).

Probabilistic output modeling complements calibration by targeting the shape and dependence structure of the predictive distribution, not only its coverage. Its main benefit is richer uncertainty information, including epistemic--aleatoric decomposition and coherent multi-horizon sample paths. The key challenge is to estimate marginal distributions, cross-horizon dependence, and uncertainty decompositions from limited target-domain evidence.

\subsection{Anomaly Detection}
Anomaly detection methods convert TSFM outputs into a reliable anomaly detector (scores or p-values). Adaptive Conformal Anomaly Detection~\citep{adaptive_conformal} uses
predictions from pretrained foundation models to construct adaptive conformal bounds. The resulting anomaly score is directly interpretable as a false alarm rate or p-value. Complexity and Statistics Guided Anomaly
Detection~\citep{complexity_statistics} studies reconstruction-based anomaly detection with TSFMs. It identifies overgeneralization and overstationarization
as two key failure modes, and improves anomaly scores by incorporating high-frequency complexity measures and restoring statistical features.

Anomaly detection benefits from TSFM outputs when forecast errors, reconstruction residuals, or calibrated bounds can be converted into interpretable evidence of abnormal behavior. The main challenge is to separate true anomalies from distribution shift, model misspecification, and benign high-frequency variation. Future work should focus on score construction, false-alarm calibration, and adaptation rules that remain reliable when anomalies are rare, labels are limited, and normal behavior changes over time.

\subsection{Development and Limitations}
Output processing methods have evolved from simple forecast correction toward richer probabilistic and calibration-aware post-processing as shown in~\cref{fig:output-uncertainty-timeline}. One line focuses on forecast refinement, where lightweight
modules such as \(\delta\)-Adapter and RefineBridge correct the initial forecast without changing the deployed backbone. 
A second line focuses on calibrated forecasting sets, so far mainly split-conformal intervals for TSFM forecasts, while adaptive, bias-corrected, and joint-region variants remain to be evaluated on TSFMs. 
More recent work models predictive distributions more explicitly, either by decomposing uncertainty or by generating correlated sample paths across horizons. 

Despite their practicality, output processing methods have their specific limitations. First, they are constrained by the information contained in the original model output.
A post-processor can correct systematic errors, but it cannot recover signals that the backbone has never represented. 
Second, split-conformal calibration of TSFM forecasts~\citep{fm_conformal_prediction} assumes exchangeability between calibration and test windows, which temporal dependence and drift violate; the online conformal variants noted in \S7.2 adapt intervals sequentially but target coverage only as a long-run average over time. 
Third, probabilistic output methods must balance statistical fidelity with computational efficiency, especially when modeling long-horizon joint dependence.

\coloredsection{compression!90!black}{Compression and Specialization Methods}
\label{sec:compression_specialization}
Model compression and specialization constitute an efficiency-driven paradigm to adapt TSFMs for deployment in resource-constrained settings or specialized domains. The works usually operate at the architectural and parameter levels, typically following the initial training phase or during the adaptation to a downstream task. The central question is how to reduce the computational footprint and latency of a TSFM while maintaining, or even enhancing, the predictive performance on a specific set of tasks.
{We identify two subcategories by how the model is compressed or specialized: knowledge distillation and pruning and specialization.}
\begin{figure}[!h]
\centering
\resizebox{\textwidth}{!}{%
\begin{tikzpicture}[
    >=Latex,
    font=\footnotesize,
    timeline/.style={
        line width=1.1pt,
        draw=compression!85!black
    },
    tick/.style={
        line width=0.8pt,
        draw=compression!85!black
    },
    eevent/.style={
        rounded corners=6pt,
        align=left,
        text width=2.85cm,
        inner sep=3pt,
        font=\scriptsize,
        draw=compression!85!black,
        fill=compression!12
    },
    year/.style={
        font=\scriptsize\bfseries,
        text=compression!85!black
    },
    tag/.style={
        rounded corners=3pt,
        draw=compression!70!black,
        fill=compression!8,
        inner sep=1.8pt,
        font=\tiny\bfseries,
        text=compression!85!black,
        align=center
    },
    note/.style={
        font=\tiny,
        text=compression!75!black,
        align=center
    }
]


\draw[timeline, -{Stealth[length=3mm]}] (-0.8,0) -- (19.6,0);


\foreach \x in {0,3.0,6.0,9.0,12.0,15.0, 18.0} {
    \draw[tick] (\x,0) -- (\x,-0);
}


\draw[tick] (3.0,0) -- (3.0,-0.55);
\node[eevent, anchor=north] at (3.0,-0.62) {
\textbf{2025-05}\\[-1pt]
\textbf{Battery-Timer}~\citep{chan_2026_battery}\\[-1pt]
};

\draw[tick] (6.0,0) -- (6.0,0.55);
\node[eevent, anchor=south] at (6.0,0.62) {
\textbf{2025-05}\\[-1pt]
\textbf{Less-is-More}~\citep{zhao2026less}\\[-1pt]
\textit{Pruning}
};

\draw[tick] (12.0,0) -- (12.0,0.55);
\node[eevent, anchor=south] at (12.0,0.62) {
\textbf{2025-09}\\[-1pt]
\textbf{Consensus-Subspace KD}~\citep{Consensus-Subspace}\\[-1pt]
};

\draw[tick] (15.0,0) -- (15.0,-0.55);
\node[eevent, anchor=north] at (15.0,-0.62) {
%
\textbf{2025-10}\\[-1pt]
\textbf{Flow-of-Ranks}~\citep{yu2026understanding}\\[-1pt]
};

\draw[tick] (18.0,0) -- (18.0,0.55);
\node[eevent, anchor=south] at (18.0,0.62) {
\textbf{2026-01}\\[-1pt]
\textbf{DistilTS}~\citep{li2026distillingtimeseriesfoundation}\\[-1pt]
};



\end{tikzpicture}%
}
\caption{Timeline of compression and specialization methods for post-training of TSFMs.}
\label{fig:compression-specialization-timeline}
\end{figure}
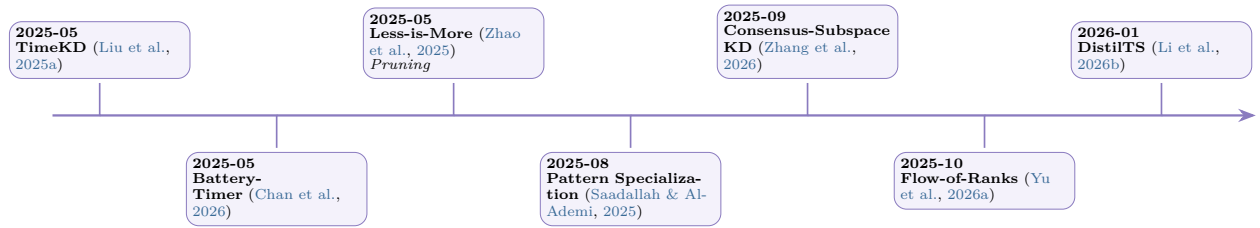
\subsection{Knowledge Distillation} 
Knowledge distillation~\citep{hinton2015distillingknowledgeneuralnetwork} compresses the knowledge of a large teacher model into a smaller student while preserving the teacher's generalization ability. Several recent works adapt this paradigm to the specific challenges of TSFMs. DistilTS~\citep{li2026distillingtimeseriesfoundation} proposes a distillation framework with a horizon-specific objective to mitigate the difficulty discrepancy between short-horizon and long-horizon forecasting, along with a temporal alignment module to bridge the architectural differences between teacher and student models. 
TimeKD~\citep{liu2025timekd}, which distills a calibrated LLM teacher with privileged ground-truth prompts into a small forecaster trained per dataset, involves no TSFM; we mention it as a transferable technique but exclude it from our catalogue. 
Battery-Timer~\citep{chan_2026_battery} proposes to first finetune the Timer TSFM with low-rank adaptation~\citep[LoRA]{hu2022lora} on battery degradation data and then apply knowledge distillation to obtain compact experts. This leads to effective and specialized models with reduced inference costs.

Knowledge distillation is attractive when a large TSFM provides strong generalization but is too costly for deployment. Its main benefit is to transfer forecasting behavior, representations, or domain-specific expertise into smaller students with lower inference cost. The main challenge is to decide what should be distilled: point forecasts, horizon-wise errors, latent representations, uncertainty estimates, or task-specific reasoning. Future work should clarify which distillation targets best preserve zero-shot generalization, calibration, and long-horizon behavior under constrained model capacity.
This echoes an open question in NLP model compression, where distilling calibrated
uncertainty and structural representations, not only output logits, remains an active research
direction~\citep{sanh2019distilbert}; TSFM distillation faces an analogous choice between
imitating point forecasts and preserving richer predictive structure.

\subsection{Pruning and Specialization} 
Pruning and specialization methods produce efficient, targeted models by removing redundant components or narrowing a general-purpose TSFM to specific patterns. \citet{zhao2026less} propose a structured pruning method to remove task-irrelevant parameters. The ``prune-then-finetune'' strategy ensures an efficient finetuning strategy to obtain specialized models with a smaller parameter space, leading to better performance than directly finetuning the original model. 
\citet{Consensus-Subspace} retain only the teacher layers with the largest drop-layer marginal loss and add trainable low-rank increments to their MLP output weights while freezing the original weights; the increments are trained to match the teacher's mean and covariance within a low-rank consensus subspace, with frequency-domain uncertainty injection to mitigate bias from the estimation subset. Although framed as distillation, the student is a layer-pruned copy of the teacher, so we discuss it here and note the teacher-alignment objective as its secondary component.
\citet{saadallah2025adaptivefinetuningpatternspecialization} fine-tune specialized copies of a forecaster on clustered patterns, but their base models are trained from scratch on each series rather than being TSFMs; we mention this as a transferable technique but exclude it from our catalogue. In~\citet{yu2026understanding}, the authors study the rank structure of time series transformers, finding that data is concentrated in low-rank subspaces which allow an efficient compression of attention layers. Building on the notion of \emph{flow-of-ranks}, which describes the mechanism by which nonlinearity increases the rank across the transformer depth, the authors propose a novel compression method. Applied post hoc, truncated SVD compresses the attention matrices of a pretrained Chronos~\citep{chronos} model to about $24\%$ of their original size without loss of accuracy; the authors' larger time and memory savings rely on a low-rank model pretrained from scratch, which falls outside post-training.

Unlike distillation, which transfers behavior from a teacher to a smaller student, pruning and specialization reduce or reshape the TSFM itself. This can preserve more of the original model while lowering inference and memory costs, but it risks removing capacity needed for transfer, calibration, or rare regimes. Future work should evaluate pruning against distillation under matched compute budgets and stress-test compressed models across horizons, domains, and regime shifts.

\subsection{Development and Limitations} 
Compression and specialization methods have developed along two main directions, as shown in~\cref{fig:compression-specialization-timeline}. The first one is largely inspired by knowledge distillation with methods using a teacher-student framework, such as DistilTS. The goal is to train compact and smaller student models to achieve the behavior of larger teachers. 
The other direction works at the architectural level and aims to retain only the components relevant to a specific task. 
This can be done by pruning~\citep{zhao2026less,Consensus-Subspace} or low-rank factorization~\citep{yu2026understanding} of the model's weights. 
It is worth noting that the two directions are complementary. Combining specialization with distillation can help develop efficient and compact experts at deployment time~\citep{chan_2026_battery}.
Despite promising results, several challenges remain. Most surveyed methods evaluate forecasting only; \citet{Consensus-Subspace} are an exception, also covering imputation, anomaly detection, and classification. It therefore remains unclear how most of these strategies transfer to other time-series tasks. Even within forecasting, the choice of distillation objective is nontrivial. Whether to align intermediate representations or final predictions, and how to handle the varying difficulty across forecast horizons, remain open questions. Evaluation practices also need strengthening. Reported efficiency metrics are also heterogeneous: some works report parameter reduction~\citep{Consensus-Subspace}, others compression ratios of attention matrices~\citep{yu2026understanding}, so methods cannot be compared on a common deployment budget. Standardized efficiency benchmarks would enable more meaningful comparisons across methods.

\section{Future Directions}
\label{sec:future_directions}

The preceding sections show that TSFM post-training is no longer synonymous with finetuning: it includes interventions on model parameters, input contexts, model composition, output distributions, and deployed architectures. We now discuss open challenges that cut across these categories and identify future directions for making post-trained TSFMs more robust, efficient, calibrated, and deployable.

\coloredsubsection{param!90!black}{Advancing Parameter Adaptation Methods}
\label{subsec:future_parameter_adaptation}
Parameter adaptation is the most direct form of TSFM post-training: it improves downstream behavior by modifying the pretrained model itself or by adding trainable components to a frozen backbone. Future work should shift from unconstrained adaptation toward controlled adaptation, where the system decides which parameters to update, how far the adapted model may move from the pretrained backbone, and when target-domain evidence is sufficient to justify adaptation. This is crucial, because parameter updates can yield strong specialization, but they also alter internal representations and may cause forgetting, instability, or overfitting under limited and nonstationary target data.

For \textbf{full fine-tuning and continual pretraining}, the evidence is mixed in an informative way. Fine-tuning can yield large gains: dataset-wise fine-tuning of Chronos-T5 Small makes it the best-performing model on Benchmark II, surpassing larger zero-shot Chronos variants and task-specific baselines~\citep{chronos}. However, gains are not guaranteed. Process-model forecasting experiments find that LoRA and full fine-tuning can improve over zero-shot TSFMs, but the improvements are dataset-dependent, sometimes small, and may disappear on smaller or more complex datasets~\citep{yu2025time}. Future work should therefore make full-model updates conditional on evidence that lighter adaptation is insufficient. This requires diagnostics that compare full fine-tuning against PEFT, prediction-head adaptation, calibration, retrieval, and in-context adaptation under matched data and compute budgets. Recent results support this direction: FourierFT adapts Chronos-Tiny with only 2,400 trainable parameters~\citep{gupta2024beyond}, TRACE reduces forecasting-head parameters by more than 70\% while using gated LoRA-module selection~\citep{trace}, and in-context fine-tuning uses continued pretraining to teach a TSFM to exploit target examples at inference time~\citep{faw2025incontext}. Finally, adaptation objectives should reflect deployment value: decision-focused fine-tuning of Moirai improves average total daily costs by 9.45\% over prediction-focused fine-tuning in feeder optimization~\citep{decision_focused_ft}.

For \textbf{parameter-efficient adaptation}, future work should move from
task-specific adapter design toward unified and adaptive lightweight adaptation.
Existing methods introduce adapters, LoRA modules, or task heads to adapt frozen TSFMs. However, these modules are often designed for specific settings. Future methods should develop general adapter interfaces that can jointly handle cross-channel dependence and task adapters. 

For \textbf{test-time and online adaptation}, future work should move from continuous updating toward selective and safe adaptation. Existing methods adapt model components during deployment to handle nonstationarity and streaming feedback. However, updating on every test sample can be unstable when the stream contains anomalous signals. Future methods should therefore decide when adaptation should be triggered, which samples are reliable for updating, and how strongly
the model should adapt. Safety mechanisms such as update rejection, rollback, and uncertainty filtering will be important for deploying TSFMs in real streaming environments.

\coloredsubsection{context!90!black}{Advancing Context Augmentation Methods}
\label{subsec:future_context_augmentation}

This family of methods improves a frozen or lightly adapted TSFM by constructing an auxiliary context
for the target series. Future work should move beyond simply adding more context and focus on reliable context construction. In particular, an effective context-augmentation system should determine what information is useful, when external context is harmful, how contextual knowledge can be stored and updated efficiently, and how the input itself should be transformed to match the distribution expected by the pretrained model. 

For \textbf{retrieval augmentation}, future work should move from similarity retrieval toward utility-aware retrieval. Existing methods retrieve related series or historical segments and inject them into the forecasting or detection process. However, temporal similarity does not always imply downstream usefulness. Two series may share similar shapes, seasonality, or frequency patterns while having different future dynamics. Therefore, retrievers should be trained or evaluated according to their contribution to the final task loss rather than only their similarity. Moreover, future methods should explicitly handle harmful retrievals by estimating retrieval confidence, filtering irrelevant series, or abstaining when no reliable context is available. Finally, retrieval-augmented TSFMs should be evaluated under realistic deployment costs, including retrieval latency, memory footprint, and end-to-end inference time.
This mirrors a parallel shift in NLP retrieval-augmented generation, where learned,
utility-aware retrievers improved over pure similarity-based retrieval~\citep{lewis2020rag};
analogous learned-retriever objectives may transfer directly to the time-series setting.

For \textbf{memory augmentation}, future work should focus on dynamic, compact, and confidence-aware memory. Memory-based methods reduce the latency and scalability issues of explicit retrieval by storing domain knowledge, temporal prototypes, or normal patterns in a compact form. The main challenge is the trade-off between memory efficiency and information fidelity. A memory module should be small enough for fast inference, but faithful enough to preserve rare events. Future methods should therefore study how to update memory under distribution shift while avoiding collapse from anomalous, noisy, or low-confidence observations. This is especially important for anomaly detection, where storing abnormal samples as normal prototypes can degrade the detector.
As summarized in Table~\ref{tab:data-regime}, memory augmentation's chief exposure is
noise sensitivity rather than shift magnitude, motivating confidence gating as the priority
safeguard over shift-adaptive retraining. A promising direction is to equip memory modules with confidence gates, so that the model can decide how strongly to rely on memory for each input. In privacy-sensitive domains such as
healthcare, another important direction is privacy-preserving memory, where raw historical series are replaced by synthetic summaries or memory representations.

For \textbf{context transformation}, future work should treat adaptation as learned context engineering. Future methods should learn
how to normalize, rescale, denoise, decompose, or re-represent the input series so that it better matches the pretraining distribution of the backbone. At the same time, such transformations should be constrained and uncertainty-aware, because overly aggressive transformations may remove informative signals. A further direction is to integrate context transformation with retrieval and memory: transformed inputs may improve retrieval quality, while retrieved examples or memory prototypes may guide how the input should be transformed.

\coloredsubsection{composition!90!black}{Advancing Model Composition Methods}
\label{subsec:future_model_composition}

This family of methods improves TSFMs by maintaining a pool of models and learning how to select or weight them. Future work should focus on reliable model allocation. In particular, an
effective model composition system should determine which models are useful for
a given input, when multiple models should be combined, and whether the routing decision is supported by meaningful temporal properties. This requires model composition to become uncertainty-aware and interpretable.

For \textbf{model selection}, future work should move from a zero-shot model
selection toward adaptation-aware selection. Existing methods select one model or a small subset of models from a candidate pool according to the target series, dataset, or task. However, the model with the best zero-shot performance is not necessarily the model that will perform best after post-training. Therefore, future selectors should estimate not only immediate performance, but also the expected gain after fine-tuning or calibration. Another important direction is dynamic model zoo maintenance. Since new TSFMs are continuously released, future systems should be able to add new models, characterize their strengths, and remove redundant models without retraining the whole router.

For \textbf{adaptive fusion}, future work should move from sample-level weighted averaging toward uncertainty-aware fusion. Existing methods combine predictions from multiple TSFMs using weights. However, a single weight for each model may be too coarse, because different models may be useful for different horizons, channels, and quantile levels. Future fusion methods should therefore learn horizon-wise, channel-wise weights. They should also combine predictive distributions rather than only point forecasts, so that uncertainty information from different models can be preserved and calibrated.
Such input-conditional weighting parallels mixture-of-experts routing developed for large-scale language and vision models~\citep{shazeer2017moe,fedus2022switch}, where sparse, learned expert gating offers a template for interpretable, input-dependent model allocation that
TSFM composition methods could adapt.

For \textbf{sequential and agentic fusion}, interpretability should also become more rigorous.
Natural language explanations are not sufficient unless they are grounded in measurable temporal properties, such as seasonality, volatility, regime shifts, or historical expert reliability. 

\coloredsubsection{output!90!black}{Advancing Output Processing and Uncertainty Control Methods}
\label{subsec:future_output_processing}

This family of methods improves TSFMs after an initial prediction. Future work should focus on
turning raw TSFM outputs into a reliable output for deployment. In particular, a reliable output processing system should determine which part of the initial output should be corrected, and how output signals can be converted into actionable evidence for downstream decisions.

For \textbf{forecast refinement}, future work should move from generic residual correction toward structured error correction. Existing methods usually refine the initial forecast by learning a map on top of the frozen backbone output. However, forecasting errors may arise from different sources,
including horizon bias, channel mismatch, seasonal misalignment, or long-term trend errors. Future refinement methods should therefore identify and correct these error components separately. 

For \textbf{forecast calibration}, future work should move from marginal interval calibration toward calibration under temporal dependence and distribution shift. Since time series are sequential and often nonstationary, coverage estimated on a fixed calibration set may fail after a distribution shift.
Future methods should adapt prediction sets over time and evaluate coverage across horizons, channels, and high-risk periods.

For \textbf{probabilistic output modeling}, future work should move from marginal predictive distributions toward coherent trajectory distributions. Future methods should preserve temporal and cross-channel dependence while remaining scalable for long horizons and high-dimensional multivariate series.

For \textbf{anomaly detection}, future work should move toward calibrated anomaly evidence. Existing methods convert forecasts or reconstructions into anomaly scores, but raw reconstruction or forecasting errors are often difficult to interpret across datasets. Future methods should produce anomaly evidence with operational meaning, such as p-values or event-level abnormality measures. 

\coloredsubsection{compression!90!black}{Advancing Compression and Specialization Methods}
\label{subsec:future_compression_specialization}

This family of methods improves TSFMs by reducing their computational cost or by specializing them for a target domain. Future work should move from simple model size reduction toward deployment-aware specialization. In particular, a reliable compression and specialization system should determine what temporal knowledge should be preserved, which computations can be safely removed, and how efficiency gains translate into practical improvements in latency, memory usage, and robustness under target-domain shift. 

For \textbf{knowledge distillation}, future work should move from output imitation toward structure- and uncertainty-aware distillation. Existing methods usually train a compact student model to imitate a larger TSFM. However, matching point forecasts alone may discard cross-horizon dependence or predictive uncertainty. Future distillation methods should therefore preserve their temporal representations, trajectory distributions, and calibrated uncertainty. 

For \textbf{pruning and structural specialization}, future work should move from static parameter removal toward adaptive and task-conditioned computation. Time series tasks often differ substantially in horizon, frequency, and noise level. A fixed pruned structure may therefore be efficient for one setting but suboptimal for another. Table~\ref{tab:data-regime} rates compression's shift-magnitude sensitivity as
medium — future task-conditioned pruning should be validated specifically against this axis
rather than series length or noise, which affect it less. Future methods should develop task-conditioned pruning. Such methods should also analyze whether the removed components correspond to redundant computation or to temporal knowledge that is only useful under other tasks.
This direction parallels task-conditioned and structured pruning explored in computer
vision and NLP~\citep{frankle2019lottery,han2016deepcompression}, where lottery-ticket-style and
magnitude-based criteria identified which parameters could be removed without full retraining ---
criteria that have not yet been systematically tested for TSFM backbones.

\subsection{Toward Agentic and Multimodal TSFMs}
Beyond the five post-training families discussed above, an important future direction is extending TSFMs post-training toward agentic and multimodal settings. 
Current TSFMs are effective in modeling numerical time series. LLMs offer complementary strengths in semantic reasoning, planning, and tool use. 
Integrating other foundation models into TSFMs post-training and enhancing them remains an open challenge.

In agentic settings, an LLM can orchestrate a complete pipeline: selecting the appropriate TSFM, constructing the query, retrieving external context, interpreting TSFM outputs, and translating output into downstream actions. For example, TSFM-generated forecasts can improve LLM-based decision-making in financial trading settings~\citep{xie2026time}. ChronoSteer~\citep{chronosteer} shows that textual events can be converted into revision instructions that steer TSFM predictions, and TS-Reasoner~\citep{yu2025tsreasoner} aligns TSFM representations with LLM reasoning for time series understanding tasks. 
These works suggest that future post-training may extend beyond adapting the TSFM itself to shaping outputs that are interpretable and actionable for LLM-based agents. Interfaces between LLMs and TSFMs, such as translating textual instructions into forecast revisions or aligning temporal representations with language-model reasoning, will likely require joint post-training. More broadly, post-training may need to extend beyond predictive accuracy, taking into account the composability and controllability of TSFM outputs within agentic pipelines.

In multimodal settings, real-world time series are usually accompanied by textual descriptions, event logs, clinical notes, visualizations, or expert annotations. However, most TSFMs are still applied and post-trained on numerical inputs alone (see~\citep{multimodaltimeseriesSurvey} for a survey on multimodal time series analysis). Recent multimodal benchmarks confirm that textual context can provide complementary information~\citep{Time-MMD, williams2025context}, for instance via question-answering~\citep{xie2026arfbench}, and early work explores unified interfaces between time series and understanding over temporal data~\citep{chattime}. 
{Effective multimodal post-training will require objectives that guide TSFMs to leverage heterogeneous contexts that they were not pretrained on, as well as calibration strategies that account for the varying informativeness of multimodal inputs across domains and time. 
This leads to post-training methods that go beyond numerical adaptation toward cross-modal alignment and selective context integration.}

\subsection{Limitations of This Survey}
This survey has four limitations worth stating explicitly. First, the evidence base is
preprint-heavy and evolves quickly: many methods discussed here are recent arXiv preprints
without peer review, and our search cutoff of April 2026 (\S1) means later work is not
reflected. Second, assignment of methods to the five taxonomy categories, and to subcategories
within them, involved subjective judgment where a method's primary locus of intervention was
not immediately clear from its stated contribution; \S2 gives our operational criteria, but
borderline cases remain possible. Third, the surveyed literature is dominated by forecasting
applications; classification, anomaly detection, imputation, and other tasks are comparatively
underrepresented across all five families, which may understate open challenges specific to
those tasks. Fourth, we discuss representative methods within each subcategory rather than
exhaustively enumerating every paper meeting our inclusion criteria (\S1), so absolute counts
of methods per category should not be read as a precise measure of research activity in that
direction.

\subsection{Deployment Risks and Broader Impact}

Post-training can introduce additional deployment risks. Under distribution
shift, adaptation or recalibration may fail silently, leading to
overconfident forecasts, unreliable prediction intervals, or misleading
anomaly scores. Retrieval and persistent memory may expose sensitive
temporal records or introduce irrelevant and contaminated context, while
model-composition pipelines may propagate errors across multiple components.
These risks are particularly consequential in healthcare, finance, and
industrial monitoring. Improvements in predictive performance, calibration,
or efficiency should therefore be accompanied by domain-specific validation,
privacy safeguards, continuous monitoring, and appropriate human oversight.
In this work, reliable deployment is a research objective concerning
empirical robustness and calibration, rather than a formal guarantee of
safety.

\section{Conclusion}
Post-training is becoming an increasingly important step in turning time series
foundation models from general-purpose pretrained backbones into reliable models for downstream
use. In our work, we analyzed this emerging area through a unifying lens covering parameter adaptation, context augmentation, model composition, output processing, uncertainty control, compression, and specialization. Looking forward, the field should move from generic performance improvement toward reliable, uncertainty-aware, and deployment-adaptive methods. We hope this study provides a useful structure for understanding current progress and for guiding future research on robust and efficient time series foundation models.

\bibliography{references}
\bibliographystyle{template/tmlr}
\appendix

\section{Auditable Method Comparison}
\label{app:comparison-table}

Table~\ref{tab:comparison} lists all surveyed methods with the attributes needed for
auditability: backbone architecture, downstream task and supervision, trained components,
evaluation setting and publication venue, and artifact availability. Category bands follow
the color coding of Fig.~\ref{fig:tsfm_posttraining_taxonomy} (A: parameter adaptation,
B: context augmentation, C: model composition, D: output processing, E: compression).

\paragraph{Verification procedure.} Each entry was verified against its primary source (full text, proceedings or publisher page, and official code repository) for (i) use of a TSFM as defined in Section~2, (ii) the post-training criterion, (iii) subcategory placement, and (iv) every table cell; every sentence of the main text that cites the entry was checked against the same source. Methods evaluated only with task-specific models are excluded from this table and mentioned in the main text only as related techniques; boundary cases are flagged in the corresponding subsection. Backbones in parentheses with ``+'' are secondary; ``In-domain'' denotes training or calibration and testing on splits of the same target dataset; the literature cutoff is April 2026, while venue and code status were updated at the time of this revision.

\rowcolors{2}{white}{gray!6}
\begin{longtable}{p{2.1cm}p{1.9cm}p{2.7cm}p{2.9cm}p{2.3cm}p{1.1cm}}
\caption{Auditable comparison of all surveyed methods.}
\label{tab:comparison}\\
\toprule
\rowcolor{white}
\textbf{Method} & \textbf{Backbone} & \textbf{Task (supervision)} & \textbf{Trained components} & \textbf{Venue (eval.\ setting)} & \textbf{Code} \\
\midrule
\endfirsthead
\toprule
\rowcolor{white}
\textbf{Method} & \textbf{Backbone} & \textbf{Task (supervision)} & \textbf{Trained components} & \textbf{Venue (eval.\ setting)} & \textbf{Code} \\
\midrule
\endhead
\bottomrule
\endfoot

\rowcolor{catA}
\multicolumn{6}{l}{\textcolor{white}{\textbf{A1. Full-parameter FT \& continual pretraining (\S4.1)}}} \\
Wass-FT & Chronos-T5 (Small) & Forecasting (labeled) & Full backbone (Wasserstein loss) & NeurIPS-W '24 (In-domain/FT) & \checkmark \\
Fin-FT & TimesFM & Forecasting (labeled) & Full backbone & IEEE CiFer '25 (In-domain/FT) & \checkmark \\
Med-FedFT & Chronos-Tiny & Forecasting (federated) & Model weights (federated; subset not stated) & Preprint '25 (Cross-site/Fed) & \textbf{--} \\
VisionTS++ & MAE (ImageNet) & Forecasting (unlabeled) & Full backbone + quantile heads & Preprint '25 (Zero-shot) & \checkmark \\

\rowcolor{catA}
\multicolumn{6}{l}{\textcolor{white}{\textbf{A2. Parameter-efficient adaptation (\S4.2)}}} \\
ChronosX & Chronos (+TimesFM, MOMENT) & Exogenous cov.\ (labeled) & Adapters/exogenous module & AISTATS '25 (In-domain/FT) & \checkmark \\
UniCA & Chronos-Bolt, TimesFM, Time-MoE & Exogenous cov.\ (labeled) & Adapter modules & ICLR '26 (In-domain) & \checkmark \\
CoRA-Cov & Sundial (+TimesFM, Chronos-Bolt) & Exogenous cov.\ (labeled) & Adapter modules & Preprint '25 (In-domain, full/few-shot) & \textbf{--} \\
WindPrompt & Timer & Forecasting (labeled) & Time-series prompts & Applied Energy '25 (In-domain/FT) & \textbf{--} \\
CoRA & MOMENT, Timer, TTM (+LLM-based) & Multivariate (labeled) & Correlation-aware adapter (+backbone in official code) & ICLR '26 (In-domain, 5\% few-shot) & \checkmark \\
AdaPTS & MOMENT & Multivariate (labeled) & Latent projection & ICML '25 (In-domain/FT) & \checkmark \\
DualWeaver & Sundial, Timer-XL & Multivariate (labeled) & Surrogate-series fusion & Preprint '26 (In-domain) & \checkmark \\
Generalized Prompt Tuning & MOMENT (+GPT4TS) & Multivariate, healthcare (labeled) & Trainable prompts & ML4H '24 (In-domain, limited data) & \checkmark \\
MixFT & Chronos-Bolt, Moirai & Forecasting (labeled) & Per-sub-domain LoRA modules & Preprint '26 (Zero-shot after FT on related data) & \textbf{--} \\
TRACE & MOMENT & Forecasting, anomaly (labeled) & PEFT modules & Neurocomputing '26 (In-domain FT; cross-dataset short-term) & \textbf{--} \\
STAR & DADA, UniTS, MOMENT, Timer & Anomaly det.\ (unsupervised) & State-aware adapter & Preprint '25 (In-domain, 10\% data) & \textbf{--} \\
MSFT & Moirai, MOMENT, UniTS & Forecasting\ (labeled) & Multi-scale adapters & NeurIPS '25 (In-domain/FT) & \checkmark \\
ULoRA-MoE & UniTS (+ GPT4TS) & Anomaly det.\ (unsupervised) & LoRA MoE experts + Gumbel-Softmax router & ICLR '25 subm., withdrawn (In-domain/FT) & -- \\
F2A & TTM, TSPulse, MOMENT & Anomaly prediction (labeled) & Decoder + task heads (encoder frozen) & Preprint '25 (In-domain + held-out) & \textbf{--} \\
Decision-FT & Moirai & Decision/optim.\ (labeled) & LoRA/DoRA adapters & Energy \& AI '25 (In-domain/FT) & \checkmark \\
FORMED & TimesFM & Classification, medical (labeled) & Shared decoding attention + label queries & ICLR '26 (Cross-domain) & \checkmark \\
VaR-Forecast & TimesFM & Forecasting (labeled) & Linear probing (transformer frozen) & Preprint '25 (In-domain; zero-shot vs FT) & \checkmark \\
RV-Forecast & TimesFM & Forecasting (labeled) & Linear probing (transformer frozen), incremental & Preprint '25 (In-domain/FT) & \textbf{--} \\

\rowcolor{catA}
\multicolumn{6}{l}{\textcolor{white}{\textbf{A3. Test-time \& online adaptation (\S4.3)}}} \\
ELF & Chronos, Moirai, TimesFM, TTM, VisionTS & Forecasting (delayed ground truth) & Lightweight online heads & ICML '25 (TTA/Online) & \textbf{--} \\
TAFAS & Chronos (one exp.) + per-dataset forecasters & Forecasting (partially observed GT) & Input/output gated calibration modules & AAAI '25 (TTA) & \checkmark \\

\rowcolor{catB}
\multicolumn{6}{l}{\textcolor{white}{\textbf{B1. Retrieval augmentation (\S5.1)}}} \\
RAF & Chronos (+Moirai, TimesFM, Lag-Llama) & Forecasting (retrieval) & None (Naive) / full backbone FT (Advanced) & Preprint '24 (Zero-shot / per-dataset FT) & \checkmark \\
TimeRAF & TTM & Forecasting (retrieval) & Learnable retriever + channel prompting (TTM frozen) & IEEE TKDE '25 (Zero-shot) & \textbf{--} \\
TS-RAG & Chronos-Bolt (+MOMENT) & Forecasting (retrieval) & Adaptive Retrieval Mixer + projector (retriever frozen) & NeurIPS '25 (Zero-shot) & \checkmark \\
Cross-RAG & Chronos-Bolt, MOMENT & Forecasting (retrieval) & Cross-attention RAG adapter & KDD '26 MiLeTS Wksp (Zero-shot) & \checkmark \\
RATFM & Time-MoE, MOMENT & Anomaly det.\ (retrieval) & TSFM fine-tuned on retrieval-augmented inputs & Preprint '25 (Leave-one-domain-out) & \checkmark \\
RAG4CTS & Chronos-2 & Exogenous cov.\ (retrieval) & None (training-free) & Preprint '26 (Zero-shot, single industrial dataset) & \textbf{--} \\

\rowcolor{catB}
\multicolumn{6}{l}{\textcolor{white}{\textbf{B2. Memory augmentation (\S5.2)}}} \\
MEMTS & Chronos-Bolt, MOMENT, Moirai, Sundial & Forecasting (labeled) & Latent memory prototypes (backbone frozen) & Preprint '26 (In-domain) & \textbf{--} \\
TS-Memory & Chronos-Bolt, Chronos-2, Sundial, TimesFM & Forecasting (mixed) & Plug-and-play memory module & KDD '26 (In-domain; cross-domain study) & \checkmark \\
MOMEMTO & MOMENT-large & Anomaly det.\ (unlabeled) & Encoder + decoder + patch memory (joint FT) & Preprint '26 (Multi-domain/few-shot/zero-shot) & \textbf{--} \\

\rowcolor{catB}
\multicolumn{6}{l}{\textcolor{white}{\textbf{B3. Context transformation (\S5.3)}}} \\
TATO & Timer, Moirai, Chronos & Forecasting (target history; search) & None (TPE-searched pipeline) & ICLR '26 (In-domain; frozen backbone) & \checkmark \\

\rowcolor{catC}
\multicolumn{6}{l}{\textcolor{white}{\textbf{C1. Model selection (\S6.1)}}} \\
Chroma & Chronos-Bolt-arch. portfolio & Forecasting (selection on target validation window) & Specialists finetuned from a shared base; no trained combiner & ICLR '26 (Zero-shot) & \textbf{--} \\
ZooCast & TSFM model zoo & Forecasting (unlabeled) & Co-embedding extractor (TSFMs frozen) & Preprint '25 (Zero-shot) & \textbf{--} \\
TimeTic & TSFM model zoo & Model selection/transferability & None (in-context TabPFN estimator) & Preprint '25 (Leave-one-dataset-out; post-FT performance) & \textbf{--} \\

\rowcolor{catC}
\multicolumn{6}{l}{\textcolor{white}{\textbf{C2. Adaptive fusion (\S6.2)}}} \\
Synapse & Sundial, Toto, Moirai & Forecasting (weights from context CRPS) & None (rule-based inverse-CRPS weighting) & TMLR '26 (Zero-shot, GIFT-Eval) & \textbf{--} \\
Boosting (Modi \& Pan) & Lag-Llama, TimeGPT & Forecasting (labeled target history) & Stacking regression + AutoGluon residual model & Preprint '25 (Single dataset, in-domain) & \checkmark \\

\rowcolor{catC}
\multicolumn{6}{l}{\textcolor{white}{\textbf{C3. Sequential \& agentic fusion (\S6.3)}}} \\
SeqFusion & PatchTST zoo; TSFM zoo (one exp.) & Forecasting (unlabeled) & Representation extractor (similarity-based selection) & Preprint '25 (Zero-shot) & \checkmark \\
Conversational TSFM & Moirai-2, Sundial, Toto, TabPFN-TS + Qwen-2.5-3B & Explainable forecasting (text-aligned) & LLM judge (SFT + GRPO; TSFMs frozen) & Preprint '25 (GIFT-Eval) & \textbf{--} \\

\rowcolor{catD}
\multicolumn{6}{l}{\textcolor{white}{\textbf{D1. Forecast refinement (\S7.1)}}} \\
$\delta$-Adapter & Sundial-S, TTM-R2 (+non-TSFM forecasters) & Forecasting (labeled) & Post-processing shift modules & ICLR '26 (In-domain, frozen backbone) & \checkmark \\
RefineBridge & Chronos-L, Moirai-L, Time-MoE-B & Forecasting, financial (labeled) & Generative bridge refiner & ICASSP '26 (In-domain; TSFM frozen, refiner per asset) & \textbf{--} \\
TFMAdapter & TimesFM, Chronos, Moirai & Exogenous cov.\ (labeled) & Per-instance ridge + GP adapter (refit per query) & CIKM '25 (Zero-shot, instance-level) & \checkmark \\

\rowcolor{catD}
\multicolumn{6}{l}{\textcolor{white}{\textbf{D2. Forecast calibration (\S7.2)}}} \\
FM+CP & Lag-Llama, Chronos(-Bolt), TimesFM(2) & Forecasting (calib.\ set) & None (split-conformal quantile) & Preprint '25 (Zero-shot/Calib) & \checkmark\\

\rowcolor{catD}
\multicolumn{6}{l}{\textcolor{white}{\textbf{D3. Probabilistic output modeling (\S7.3)}}} \\
Corr.\ Sample Paths & TimesFM, Chronos-Bolt, TiRex & Forecasting (context autocorr.) & Copula post-processor (trained variant: copula network) & NeurIPS '25 Wksp (Zero-shot post-hoc) & \textbf{--} \\

\rowcolor{catD}
\multicolumn{6}{l}{\textcolor{white}{\textbf{D4. Anomaly detection (\S7.4)}}} \\
Adaptive Conformal AD & TSFM & Anomaly det.\ (calibration data) & Adaptive conformal quantile estimator & ICLR '26 (Zero-shot/Online) & \checkmark \\
Complexity+Stats AD & MOMENT; Chronos, Moirai, Time-MoE & Anomaly det.\ (unsupervised) & Recon.\ head with instance stats (encoder frozen) + complexity-weighted ensemble & ICLR '26 (Per-dataset unsupervised / zero-shot) & \textbf{--} \\

\rowcolor{catE}
\multicolumn{6}{l}{\textcolor{white}{\textbf{E1. Knowledge distillation (\S8.1)}}} \\
DistilTS & Time-MoE, Moirai, Chronos, TimesFM, MOMENT (teachers) & Forecasting (teacher-student) & Student + alignment module & ICASSP '26 (In-domain, per-dataset students) & \checkmark \\
Battery-Timer & Timer-84M (teacher) & Forecasting, battery (supervised) & LoRA on Timer + full task-specific students & RESS '26 (Zero-shot teacher; in-domain students) & \checkmark \\

\rowcolor{catE}
\multicolumn{6}{l}{\textcolor{white}{\textbf{E2. Pruning \& specialization (\S8.2)}}} \\
Less-is-More & TTM, Time-MoE, Moirai, Chronos-Bolt, TimesFM & Forecasting (pruning + FT) & Structured pruning + FT of retained weights & NeurIPS '25 (In-domain FT) & \checkmark \\
Flow-of-Ranks & Chronos & Forecasting (compression) & None (training-free truncated SVD of attention) & ICLR '26 (In-domain/Zero-shot) & \checkmark \\
Consensus-Subspace KD & MOMENT-Large & Forecast., imput., AD, classif. (teacher-student) & Low-rank increments on retained teacher layers & ICML '26 (Zero-shot/LP) & \checkmark \\

\end{longtable}

\end{document}